\documentclass{article}

\usepackage[numbers]{natbib} %

\usepackage[preprint]{neurips_2024}

\usepackage[utf8]{inputenc} 
\usepackage[T1]{fontenc}    
\usepackage{graphicx}
\usepackage{multirow}
\usepackage{subcaption}
\usepackage{hyperref}       
\usepackage{url}            
\usepackage{booktabs}       
\usepackage{amsfonts}       
\usepackage{nicefrac}       
\usepackage{microtype}      
\usepackage{xcolor}         
\usepackage{amsmath}
\usepackage{caption}

\title{MORE-CLEAR: Multimodal Offline Reinforcement learning for Clinical notes Leveraged Enhanced State Representation}

\def \institution{Seoul National University Hospital }
\def \irbNO{IRB-000-000-000}
\def \country{Republic of Korea }

\author{
\textbf{Yooseok Lim}\textsuperscript{\textbf{1}} \thanks{These authors contributed equally.} \quad
\textbf{ByoungJun Jeon}\textsuperscript{\textbf{1}} \footnotemark[1] \quad
\textbf{Seong-A Park}\textsuperscript{\textbf{1}} \quad
\textbf{Jisoo Lee}\textsuperscript{\textbf{3}} \quad
\textbf{Sae Won Choi}\textsuperscript{\textbf{2}} \quad \\[1pt]
\textbf{Chang Wook Jeong}\textsuperscript{\textbf{1,3}} \quad
\textbf{Ho-Geol Ryu}\textsuperscript{\textbf{1}} \quad
\textbf{Hongyeol Lee}\textsuperscript{\textbf{1}} \footnotemark[2] \quad
\textbf{Hyun-Lim Yang}\textsuperscript{\textbf{1,3}} \thanks{
  Corresponding author. \\
  \texttt{takumama@naver.com, hlyang@snu.ac.kr}
} 
\\[2pt]
\textsuperscript{1}Seoul National University Hospital \quad
\textsuperscript{2}Bucheon Sejong Hospital \quad \\[1pt]
\textsuperscript{3}Seoul National University
}

\begin{document}

\maketitle

\begin{abstract}
Sepsis, a life-threatening inflammatory response to infection, causes organ dysfunction, making early detection and optimal management critical.
Previous reinforcement learning (RL) approaches to sepsis management rely primarily on structured data, such as lab results or vital signs, and on a dearth of a comprehensive understanding of the patient's condition.
In this work, we propose a Multimodal Offline REinforcement learning for Clinical notes Leveraged Enhanced stAte Representation (MORE-CLEAR) framework for sepsis control in intensive care units.
MORE-CLEAR employs pre-trained large-scale language models (LLMs) to facilitate the extraction of rich semantic representations from clinical notes, preserving clinical context and improving patient state representation.
Gated fusion and cross-modal attention allow dynamic weight adjustment in the context of time and the effective integration of multimodal data.
Extensive cross-validation using two public (MIMIC-III and MIMIC-IV) and one private dataset demonstrates that MORE-CLEAR significantly improves estimated survival rate and policy performance compared to single-modal RL approaches.
To our knowledge, this is the first to leverage LLM capabilities within a multimodal offline RL for better state representation in medical applications.
This approach can potentially expedite the treatment and management of sepsis by enabling reinforcement learning models to propose enhanced actions based on a more comprehensive understanding of patient conditions.
\end{abstract}

\section{Introduction}
Sepsis, characterized by a dysregulated host response to infection, causes organ dysfunction with a high mortality risk. 
It remains the leading cause of death in critically ill patients \cite{1:singer:2016,2:vincent:2014}.
Early identification of sepsis and prompt intervention in the intensive care unit (ICU) are pivotal for improving patient outcomes, highlighting the importance of developing effective therapeutic strategies.
However, timely recognition of sepsis is challenging due to the multitude of potential etiologies and the rapid progression of the disease \cite{3:Russell:2006,4:martin:2003}.
Furthermore, inotropes and fluid therapy can be administered to resuscitate patients from dangerously low hypotensive condition, however, the appropriate timing and amount of these treatments remain controversial \cite{sepsis_timing_1, sepsis_timing_2, sepsis_timing_3, fluid_volume_1, fluid_volume_2}.   

In recent years, reinforcement learning (RL) has emerged as a promising approach for identifying optimal treatment strategies in complex and uncertain clinical environments \cite{RL_medicine_1, RL_medicine_2, RL_medicine_review}. 
By formulating the patient care as a sequential decision-making problem---comprising states, actions, and rewards---RL facilitates the development of data-driven, personalized treatment policies. 
RL research in sepsis management relies predominantly on structured data, such as laboratory testing (lab) results and vital signs \cite{5:komor:2018,6:choi:2024,7:wang:2024,8:liang:2025}.
While these data are valuable from a clinical perspective, they are often characterized by substantial missingness, noise, and irregular sampling, which collectively undermine the reliability of state representations and hinder effective policy learning \cite{9:nauka:2025}.
Although recent approaches have sought to alleviate sparsity and bias in electronic medical records (EMRs) \cite{11:zhu:2024, 12:olaimat:2024}, a fundamental limitation remains in that structured data alone is insufficient for capturing the complexity of patient states \cite{13:shukla:2020, 14:ruan:2025}. 

In clinical practice, decisions regarding treatment planning and medication administration are often informed by clinical notes, which provide nuanced contextual records, including medical history, symptom evolution, and response to interventions \cite{15:raghavan:2014, 16:mathiou:2016}. 
Incorporating clinical notes can compensate for structured data limitations and enhance patient state representations. 
Although prior research in the domain of supervised learning has evidenced the efficacy of such multimodal integration \cite{17:teles:2025}, the potential of this approach within RL remains under-explored. 
In particular, given the irregular documentation of clinical notes during events and the imbalanced distribution of information over time, effective integration of multimodal clinical records that accounts for these characteristics poses a significant challenge, one that must be surmounted to facilitate robust policy learning.

To address these challenges, we propose MORE-CLEAR (Multimodal Offline REinforcement learning for Clinical notes Leveraged Enhanced stAte Representation), a novel multimodal offline RL framework tailored for sepsis management in the ICU. 
MORE-CLEAR employs a pre-trained large language model (LLM) to extract enriched embedding vectors containing semantic representations from clinical notes. 
Motivated by clinical reasoning, in which early patient information guides understanding prior treatment trajectories, we model the initial-time note as a context vector injected at each decision to mitigate information sparsity and preserve temporal coherence. 
A gated fusion mechanism is introduced to facilitate effective integration of the context vectors and observation vectors.
Embeddings for structured data, i.e., lab results and vital signs, are obtained using a refined multi-layer perception (MLP)-based encoder to handle missing values and complex feature interactions.
A bidirectional cross-modal attention mechanism is then employed to integrate salient information from structured and unstructured modalities, forming the final state representation for RL. 
The effectiveness of MORE-CLEAR was evaluated on two publicly available datasets, MIMIC-III and MIMIC-IV, and a private ICU dataset. 
Results demonstrate that incorporating clinical notes through MORE-CLEAR leads to substantial improvements in survival rate estimation and policy performance compared to unimodal RL approaches. 

Extracting comprehensive patient state representations for robust RL remains a fundamental challenge in the medical field. MORE-CLEAR addresses this issue through the following key contributions:
(i) We show that integrating clinical notes with lab results using LLMs significantly enhances policy performance compared to unimodal baselines.
(ii) We propose a context-aware gated fusion that encodes the initial patient information as a persistent context vector and fuses it with temporal observations to enhance RL performance.
(iii) A bidirectional cross-modal attention mechanism integrating clinical notes and lab tests is shown to yield a statistically significant improvement in policy performance.
(iv) Extensive cross-evaluations on MIMIC-III, MIMIC-IV, and a private dataset demonstrate that MORE-CLEAR consistently outperforms unimodal frameworks in off-policy evaluation (OPE) metrics and predicted survival rates based on enhanced state representation.

\section{Related Work}
\subsection{Sepsis Treatment Optimization via RL}
Prior work on RL for sepsis management has primarily focused on optimizing vasopressor and intravenous fluid dosages \cite{5:komor:2018, 6:choi:2024, 8:liang:2025, 18:tu:2025,19:kaushik:2022,20:fang:2024}, or devising strategies for antibiotic administration \cite{7:wang:2024} using conventional off-policy algorithms, such as Deep Q-Networks (DQN) and Double DQN (DDQN). 
Although many studies have incorporated rule-based constraints in their training to ensure reliability in real-world clinical settings \cite{7:wang:2024,20:fang:2024}, the fundamental limitation of relying solely on structured data hinders capturing contextual factors, such as patient comorbidities and treatment history.
Additionally, the robustness and efficacy of the model in various clinical settings and cohorts have not been thoroughly evaluated, as only one-way evaluations have been considered.

\subsection{Offline RL}
Offline RL provides a practical framework for policy learning in data-constrained domains such as healthcare, autonomous driving, and robotics \cite{20.5:kiran:2021, 20.7:singh:2022}. 
Recently, it has been explored for treating sepsis to mitigate distributional shifts between the learned and behavioral policies \cite{18:tu:2025,19:kaushik:2022,20:fang:2024}.
A key challenge in Offline RL is reducing the overestimation of Q-value for out-of-distribution (OOD) actions compared to the behavior policy.
Although various methods, such as Batch-Constrained Q-Learning (BCQ) \cite{21:fjjimoto:2019}, Bootstrapping error accumulation reduction \cite{22:kumar:2019}, Conservative Q-Learning (CQL) \cite{23:kumar:2020}, or Implicit Q-Learning \cite{24:kostrikov:2021}, have been proposed to address this challenge, a one-size-fits-all solution for all scenarios has yet to be found.
In the medical domain, several ideas, such as variations of BCQ \cite{25:killian:2020}, critical patient state \cite{26:fatemi:2022}, and sampling strategies \cite{27:nambiar:2023}, have been proposed to address the problem; however, investigating the influence of multimodal state representations on policy performance remains under-explored.

\subsection{Multimodal fusion in Medicine}
Multimodal learning has garnered increasing attention in the medical domain, aiming to integrate diverse data types such as EMRs, clinical notes, imaging, and genomics \cite{28:song:2021,29:li:2021,30:zhang:2023,31:adler:2022,32:kocak:2012,33:chaa:2023,34:paraskar:2025}. 
Notable contributions include image fusion studies employing multiple modalities such as MRI, PET, and CT scans \cite{28:song:2021,29:li:2021,30:zhang:2023,31:adler:2022}; analyses leveraging biosignals like EEG and ECG \cite{32:kocak:2012,33:chaa:2023}; and integrative approaches involving genomic and other omics data \cite{34:paraskar:2025}. 
Additionally, researchers have explored models that combine lab results with clinical notes to enhance predictive performance \cite{17:teles:2025}. 
Despite encouraging outcomes in predictive tasks, these methods remain confined to classifying specific data, hindering optimal policy identification in uninterrupted clinical settings.

\section{Background}
\subsection{Multimodal Markov Decision Process}
RL provides a principled framework for addressing sequential decision-making problems, typically formalized as a Markov decision process (MDP) \cite{35:sutton:1998}. 
In this study, we model the progression of patient states as an MDP with multiple modalities of observation. 
The multimodal MDP is specified by the tuple \((\mathcal{S},\mathcal{A},\mathcal{P},r,\gamma),\) with $\mathcal{S} = \prod_{i=1}^d \mathrm{\mathcal{O}^{M_i}}$, which is the joint observation space formed by $d$ modalities, where  \(\mathcal{O}^{M_i}\) denotes the observation space of the \(i\)-th modality. 
$\mathcal{A}$ denotes the set of all possible actions. 
The transition probability function $\mathcal{P} \colon \mathcal{S} \times \mathcal{A} \times \mathcal{S} \to [0,1]$ is defined such that, for any joint observation  \(o_t = \bigl(o_t^{M_1},o_t^{M_2},\dots,o_t^{M_d}\bigr)\) and action $a_t$, the probability \(P\bigl(o_{t+1}\mid o_t,a_t\bigr)\) represents the likelihood of transitioning to the next joint observation \(o_{t+1} = \bigl(o_{t+1}^{M_1},o_{t+1}^{M_2},\dots,o_{t+1}^{M_d}\bigr)\).
The function \(r:\mathcal{S}\times \mathcal{A}\to\mathbb{R}\) denotes a reward function providing the expected reward. 
\(\gamma\in[0,1)\) is the discount factor. 
The goal of an RL is to find a policy \(\pi : S \times A \to [0,1]\) that maximizes the expected cumulative discounted return \(\sum_{t=0}^{\infty} \gamma^t r_t\).

We conduct multimodal RL in an offline setting \cite{36:levine:2020}, where the policy \(\pi \) is learned from a fixed dataset \(\mathcal{D}\) collected under a behavior policy \(\pi_\beta\), without any additional online interaction.

\subsection{Q-Learning and Deep Q-Networks (DQN)}
Q-Learning \cite{37:watkins:1992} is an off-policy, model-free reinforcement learning algorithm designed to approximate the optimal action-value function \( Q^*(s,a) \) without requiring knowledge of the environment’s transition dynamics. 
Q-Learning is fundamentally grounded in the Bellman optimality equation,

\begin{equation}
Q^*(s,a) = \mathbb{E}_{s'}\bigl[r(s,a) + \gamma \max_{a'} Q^*(s',a')\bigr],
\end{equation}
which it approximates through sample-based updates. Specifically, for each observed transition \((s_t, a_t, r_t, s_{t+1})\), the Q-value is updated as follows:

\begin{equation}
Q(s_t,a_t) \leftarrow Q(s_t,a_t) + \eta \left[r_t + \gamma \max_{a'} Q(s_{t+1},a') - Q(s_t,a_t)\right],
\end{equation}
where \( \eta \in (0,1) \) is the learning rate. 
Through repeated interaction with the environment, this update rule allows the Q-values to progressively converge, ultimately enabling the agent to derive the optimal policy \( \pi^*(s) = \arg\max_a Q^*(s,a) \).

However, Q-learning becomes impractical in high-dimensional or continuous state spaces.
To address this issue, DQN utilizes a deep neural network to approximate the Q-function \( Q(s, a; \theta) \), where \( \theta \) denotes the parameters of the Q-network.
In DQN, the target \( y \) used for training is defined as:

\begin{equation}
y = r_t + \gamma \max_{a_{t+1}} Q(s_{t+1}, a_{t+1}; \theta^{-}),
\end{equation}
where \( \theta^{-} \) represents the parameters of a fixed target network from the previous iteration, and the \( \max \) operator selects the action that yields the highest Q-value in the next state. 
Based on this target, the current network parameters \( \theta\) are updated to minimize the following loss function:

\begin{equation}
L(\theta)
= \mathbb{E}_{(s_t,a_t)\sim\pi_b}\Bigl[\bigl(y - Q(s_t,a_t;\theta)\bigr)^2\Bigr].
\end{equation}

\section{Method}
\subsection{Problem Formulation}
In this section, we formalize the problem setting of multimodal RL for optimizing sepsis treatment. 
The patient state \(\mathcal{S} = O^{M_l} \times O^{M_n} \) comprises observations from two distinct modalities: structured data, i.e., lab results and vital signs, \(O^{M_l}\) and unstructured clinical notes \(O^{M_n}\). 
The action space \(\mathcal{A}\) is defined as a discrete set of treatment options, specifically representing dosage levels of vasopressors and intravenous fluids. 
Each action \(a_t \in \mathcal{A}\) denotes the treatment administered at time step \(t\). 
The reward function \(r(s_t, a_t)\) quantifies the clinical outcome of applying \(a_t\) in state \(s_t\), which follows the primary outcome of this study.
The transition function \(\mathcal{P}\) specifies the evolution of the patient’s state over time, and an episode \(\tau\) represents a sequence of clinician–patient interactions. 
The treatment-optimization objective is to identify a policy \(\pi\) that maximizes the expected discounted cumulative clinical benefit: \(\arg\max_{\pi}\;\mathbb{E}_{\tau \sim \pi}\!\left[\sum_{(s,a)\in\tau} \gamma\,r(s,a)\right]\).

\subsection{Conservative Q-Learning (CQL)}
Our framework is motivated by CQL \cite{23:kumar:2020}, an offline RL algorithm that has demonstrated strong performance in the sepsis treatment task \cite{18:tu:2025, 19:kaushik:2022, 27:nambiar:2023}. 
CQL modifies the standard Bellman objective by adding a conservative regularization term to mitigate the tendency of conventional DQN to overestimate Q-values.

\begin{equation}
\begin{split}
L_{\mathrm{CQL}}(\theta) =\;&
\tfrac12\mathbb{E}_{(s,a,r,s')\sim \mathcal{D}}\!\biggl[\Bigl(Q_\theta(s,a)-\bigl(r + \gamma\,\mathbb{E}_{a'\sim\pi}[Q_{\bar\theta}(s',a')]\bigr)\Bigr)^2\biggr] \\
&+\;\alpha\Bigl(\mathbb{E}_{s\sim\mathcal{D},\,a\sim\mu}[Q_\theta(s,a)]
\;-\;\mathbb{E}_{(s,a)\sim\mathcal{D}}[Q_\theta(s,a)]\Bigr).
\end{split}
\end{equation}
\(\mu\) denotes a distribution over actions intended to represent OOD behavior at state \(s\). 
By suppressing the Q-values of actions sampled from \(\mu\), CQL prevents the policy from overestimating the value of actions that are not well represented in the offline dataset. 
Conversely, by encouraging higher Q-values for state–action pairs observed in the dataset \(\mathcal{D}\), the learned Q-function is guided to assign sufficiently high values to actions actually taken in the data.

The hyperparameter \(\alpha > 0\) controls the strength of the conservative regularization. 
Larger \(\alpha\) intensifies the OOD penalty, promoting more rigorous suppression of unobserved action values and enhancing robustness. 
In contrast, excessively small \(\alpha\) diminishes the conservative effect, potentially reviving the overoptimistic bias commonly observed in standard DQN formulations.
Consequently, \(\alpha\) plays a crucial role in balancing OOD‐action suppression and learning stability.

\begin{figure*}[ht]
    \centering
    \includegraphics[width=1.0\linewidth]{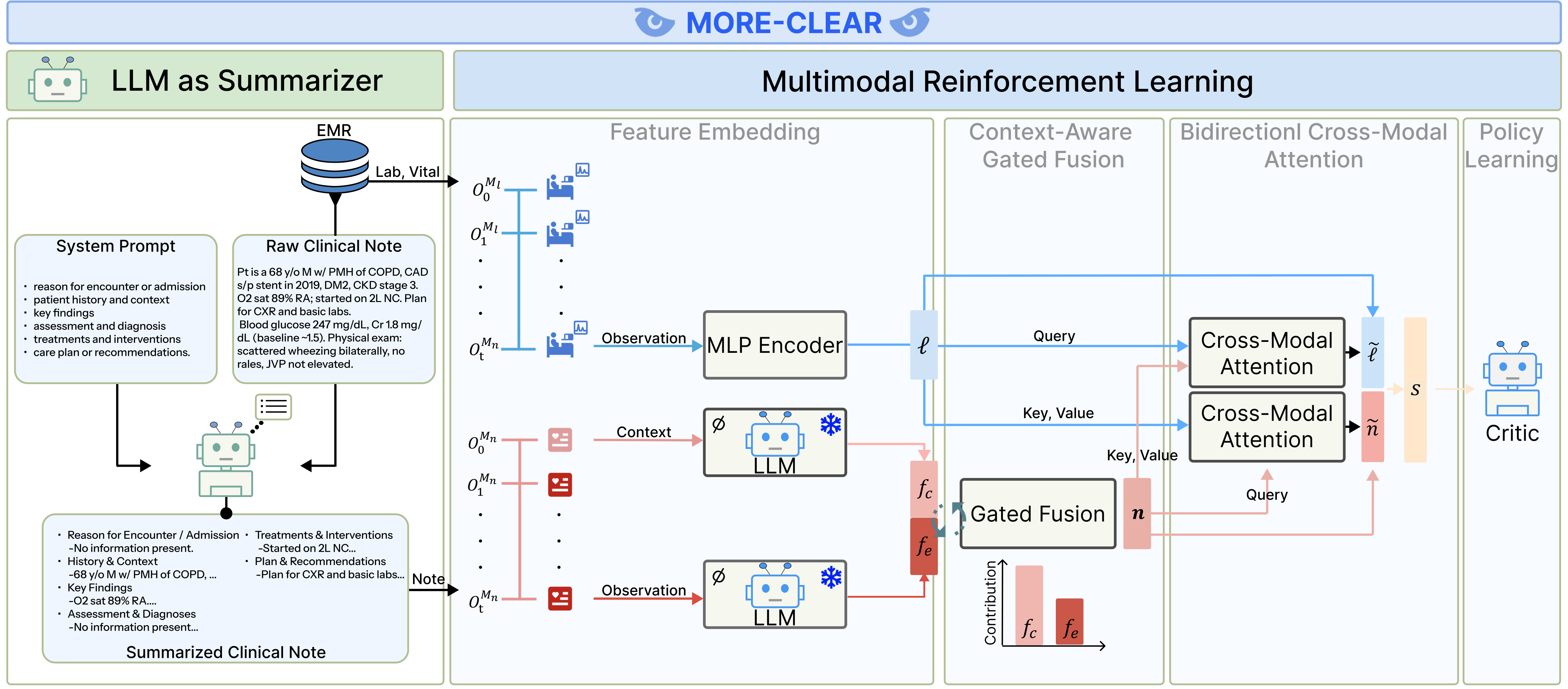}
    \caption{MORE-CLEAR framework}
    \label{fig:overview}
\end{figure*}

\subsection{MORE-CLEAR framework}
MORE-CLEAR (Figure \ref{fig:overview}) integrates structured data, i.e, lab results and vital signs, and clinical notes to derive a more enriched and comprehensive state representation, enhancing the quality of offline RL-based policy optimization.
In our framework, raw clinical notes are initially summarized using an LLM, and the generated summaries are encoded into dense vectors utilizing LLM-based encoders. 
The clinical note is further categorized into context and observation components.
Each component is independently encoded and subsequently fused using a gated fusion mechanism. 
Concurrently, structured data is processed using a data-specific MLP-based encoder motivated by MLP-Mixer \cite{tol:ranzato:2021}.
Ultimately, the bidirectional cross-modal attention module integrates embeddings from the textual and structured modalities to construct a unified state representation that facilitates robust policy learning.

\subsection{Structured Summarization with LLM for Multimodal Embedding}
Our method constructs sequential trajectories by concatenating all notes recorded within fixed time intervals.
This process often produces long sequences containing a mixture of diverse information, which complicates the identification of key content and occasionally results in inputs that exceed the token limits.
To address this, we employ an open-source LLM to summarize raw clinical notes and generate meaningful note embeddings.
A structured prompting guides LLM to summarize with organized content into clinically meaningful sections: reason for encounter or admission, patient history and context, key findings, assessment and diagnosis, treatments and interventions, and care plan or recommendations. 
Gemma-3-27B-it \cite{gemma3} is employed for the note summarization LLM according to our   
experiments in Appendix A.
Subsequent to the summarization, the structured text is fed into the LLMs, which employ average pooling on the hidden state of the final layer to extract the latent representation

\subsection{Context-Aware Gated Fusion}
We introduce a context-aware gated fusion mechanism to extract enriched patient state representations from clinical notes \(o^{M_n}\). 
Henceforth, the initial note \(o^{M_n}_{0}\) refer the context note and \(o^{M_n}_{t}\) refer as the event observation note at time \(t\).
The context note often offers implicit contextual cues that critically inform subsequent clinical decision-making, while the observation note captures local, time-specific information, including the patient's evolving condition and treatment progress.
Therefore, context-aware gated fusion is designed to maintain contextual notes to retain information about the patient's underlying medical conditions, history, presenting complaints, and the cause of hospitalization, while dynamically integrating them with event observation notes from each time frame to yield more expressive patient state representations.

In our context-aware gated fusion mechanism, \(o^{M_n}_{0}\) and \(o^{M_n}_{t}\) are projected into a high-dimensional vector space by a pre-trained LLM encoder \(\phi\). 

\begin{equation}
\begin{aligned}
f_c &= \phi\bigl(o^{M_n}_{0}\bigr) \;\in\; \mathbb{R}^{d_n}, & f_e &= \phi\bigl(o^{M_n}_{t}\bigr) \;\in\; \mathbb{R}^{d_n}.
\end{aligned}
\end{equation}
\(d_n\) denotes the dimensionality of the note embedding. 
The final note representation \(\mathbf{n}_{t} \in \mathbb{R}^d\) is computed as an adaptive, weighted combination of the context vector \(f_c\) and the event observation vector \(f_e\). 
A gating mechanism modulates this combination. 
\(\boldsymbol{\psi}_{t}\) is parameterized by a learnable weight \(W \in \mathbb{R}^{d \times 2d}\) and bias \(\mathbf{b} \in \mathbb{R}^d\).

\begin{equation}
\begin{aligned}
\boldsymbol{\psi}_{t}= \sigma\bigl(W\,[f_c; f_e] + \mathbf{b}\bigr)\;\in\;(0,1)^d, \\
\mathbf{n}_{t}= \boldsymbol{\psi}_{t} \odot f_c\;+\;(1 - \boldsymbol{\psi}_{t}) \odot f_e.
\end{aligned}
\end{equation}
Here, \(\sigma(\cdot)\) is the sigmoid function, \([\cdot\,;\,\cdot]\) denotes concatenation, and \(\odot\) is the element-wise product. 
The context-aware representation \(\mathbf{n}_t\) selectively integrates episode-level context with time-specific information, enabling enriched state representations for RL optimization.

\subsection{Bidirectional Cross-Modal Attention}
Clinical notes offer therapeutic context and insights, while also encompassing diverse content such as auxiliary procedures, clinician–patient communications, and detailed patient behaviors. 
In contrast, structured data, i.e., lab results and vital signs, provide quantitative measurements of a patient’s physiological state at specific time points, though often limited by missing values and irregular sampling.
In clinical practice, accurate patient assessment depends on the integration of these two complementary modalities \cite{38:natar:2010,39:coli:2019}. 
Motivated by this insight, we introduce a bidirectional cross-modal attention module that captures the interaction between the two modalities to selectively extract clinically meaningful information.

Let \(\mathbf{n},\ell\in\mathbb{R}^d\) denote the clinical note from gated fusion and structured data embedding vectors, respectively. 
We employ a bidirectional cross-modal attention mechanism to capture both structured data (l, v)\(\to\)note and note\(\to\)(l,v) interactions. Each modality is first projected into the query, key, and value spaces, as defined below:

\begin{align}
Q^{\mathrm{(l,v)\to note}}&=W_{\ell}^{Q}\,\ell,&
K^{\mathrm{(l,v)\to note}}&=W_{\mathbf{n}}^{K}\,\mathbf{n},&
V^{\mathrm{(l,v)\to note}}&=W_{\mathbf{n}}^{V}\,\mathbf{n},\\
Q^{\mathrm{note\to (l,v)}}&=W_{\mathbf{n}}^{Q}\,\mathbf{n},&
K^{\mathrm{note\to (l,v)}}&=W_{\ell}^{K}\,\ell,&
V^{\mathrm{note\to (l,v)}}&=W_{\ell}^{V}\,\ell,
\end{align}
where each \(W^{\cdot}\in\mathbb{R}^{d_k\times d}\) is learnable and \(d_k\) is the attention dimensionality. 
For each direction \(u\in\{\mathrm{(l,v)\to note},\,\mathrm{note\to (l,v)}\}\), attention weights \(\alpha\) and attention-weighted features \(A\) are computed as follows:

\begin{equation}
\alpha^{(u)} = \mathrm{softmax}\!\biggl(\frac{Q^{(u)}\,(K^{(u)})^\top}{\sqrt{d_k}}\biggr),
\quad
A^{(u)} = \alpha^{(u)}\,V^{(u)} \;\in\;\mathbb{R}^{1\times d}.
\end{equation}
Next, each original feature (\(\mathbf{n},\ell\)) is concatenated with its corresponding attention vector and then passed through a linear transformation.

\begin{align}
\tilde{\ell} &= W_{\ell}\,\bigl[\ell;\,A^{\mathrm{note\to (l,v)}}\bigr]\;\in\;\mathbb{R}^d,\\
\tilde{n}      &= W_{\mathbf{n}}\,\bigl[\mathbf{n};\,A^{\mathrm{(l,v)\to note}}\bigr]\;\in\;\mathbb{R}^d,
\end{align}
where \(W_{\ell},W_{\mathbf{n}}\in\mathbb{R}^{d\times2d}\). 
Finally, the fused state representation is calculated as:
\begin{equation}
s = \bigl[\tilde{\ell};\,\tilde{\mathbf{n}}\bigr]\;\in\;\mathbb{R}^{2d}.
\end{equation}

\begin{table}[ht]
  \centering
  \captionsetup{skip=10pt}
  \scriptsize
  \setlength{\tabcolsep}{4pt}
  \begin{tabular}{lccc}
      \toprule
      Variable                   & MIMIC-III (n=11,114) & MIMIC-IV (n=10,203) & Private Dataset (n=599)   \\
      \midrule
      Age (Mean $\pm$ SD)        & 64.14 $\pm$ 16.94       & 64.86 $\pm$ 16.26      & 66.59 $\pm$ 16.15\\
      Weight (kg, Mean $\pm$ SD) & 83.29 $\pm$ 24.63       & 83.13 $\pm$ 24.95      & 57.16 $\pm$ 13.40\\
      SOFA (Mean $\pm$ SD)       & 5.39 $\pm$ 3.30         & 5.16 $\pm$ 2.84        & 6.41 $\pm$ 3.35  \\
      GCS (Mean $\pm$ SD)        & 12.48 $\pm$ 3.31        & 12.95 $\pm$ 3.29       & 9.87 $\pm$ 3.35  \\
      Female (\%)                & 56                    & 58                   & 37             \\
      90-day mortality (\%)      & 22                    & 26                   & 24             \\
      \bottomrule
  \end{tabular}
  \caption{Dataset Statistics}
  \label{tab1}
\end{table}

\section{Experimental setups}
\subsection{Sepsis Treatment Optimization}
\subsubsection{Dataset and ethical approval}
We employed two publicly available datasets, MIMIC-III \cite{40:johnson:2016} and MIMIC-IV \cite{41:johnson:2023}, along with one private dataset for experiments. 
A private dataset (PD) was retrieved from the prospective registry of ICU patients at Seoul National University Hospital in the \country from April 2022 to March 2025 via the clinical data warehouse.
The collection of data and subsequent analysis were approved by the institutional review board of \institution (\irbNO), with a waiver for written informed consent due to the study’s retrospective design and data anonymity.
The inclusion and exclusion criteria were identical to those in previous research \cite{5:komor:2018,6:choi:2024}.
Table \ref{tab1} summarizes key feature statistics related to sepsis.

\subsubsection{Task}
The patient state \(s_t\) is defined at four-hour intervals and consists of two primary modalities: structured data \(O^{M_l}\), which comprises 42 clinical variables, and clinical notes \(O^{M_n}\), which encompass nursing records, physician documentation, discharge summaries, and other information.
Each episode spans from 24 hours prior to 48 hours after the suspected onset of sepsis, capturing the critical period for therapeutic intervention. 
The action \(a_t\) denotes a joint decision over intravenous (IV) fluid and vasopressor administration, each discretized into five levels, yielding 25 unique treatment combinations. 
The reward \( r_t \) is defined such that all intermediate time steps are assigned a reward of 0, whereas the terminal step (\(T\)) yields a reward of \(+1\) if the patient survives for 90 days and \(-1\) otherwise. 
The reward function is defined as

\begin{equation}
r_t =
\begin{cases}
0, & t < T, \\
+1, & \text{if the patient survives for 90 days after discharge}, \\
-1, & \text{if the patient dies},
\end{cases}
\end{equation}

\subsubsection{Preprocessing}
All datasets were preprocessed using a unified pipeline adapted from public implementations \footnote{\url{https://github.com/matthieukomorowski/AI_Clinician}} \footnote{\url{https://github.com/cmudig/AI-Clinician-MIMICIV}}, with minor modifications to ensure consistency.
Specifically, four variables from the original set of 47 clinical features \cite{5:komor:2018}---readmission, Elixhauser score, mechanical ventilation, and cumulative fluid balance---were excluded. 
The vasopressor drugs included vasopressin, norepinephrine, and dopamine. 
IV fluids included the same or similar categories as normal saline, plasma solution, albumin, and Hartmann’s solution. 
Apart from these adjustments, all preprocessing steps were consistent with the original pipeline.

\subsection{Evaluation Metrics}
A rigorous evaluation of the MORE-CLEAR's policy was conducted by employing four off-policy evaluation (OPE) metrics, including weighted importance sampling (WIS) \cite{45:mahmood:2014}, doubly robust (DR) estimator \cite{44:jiang:2016}, fitted Q-evaluation (FQE) \cite{43:gordon:1999}, and offline policy evaluation with re-weighted aggregates (OPERA) \cite{42:nie:2024}.
Beyond these quantitative measures, we further examined the policy's clinical relevance by introducing the Behavioral Discrepancy Estimated Survival Rate (BDESR), a metric designed to capture alignment between RL policy actions and clinician actions.

\subsubsection{Behavioral Discrepancy Estimated Survival Rate}
We define the BDESR as a means to estimate the clinical effectiveness of an RL policy. 
This metric quantifies the degree of discrepancy between actions administered by the RL policy and clinician actions recorded in the batch dataset, with the corresponding survival rate estimated based on behavioral inconsistency.

For each patient episode \(i\), divergences between the RL policy and clinician actions over time are computed separately for the two actions: IV fluid and vasopressor administration. 

\begin{equation}
m_{\text{iv}}^{(i)} = \frac{1}{T_i} \sum_{t=1}^{T_i} \bigl|\mathit{\hat{A}}_{\text{iv},t}^{(i)} - \mathit{A}_{\text{iv},t}^{(i)}\bigr|, 
\quad
m_{\text{vaso}}^{(i)} = \frac{1}{T_i} \sum_{t=1}^{T_i} \bigl|\mathit{\hat{A}}_{\text{vaso},t}^{(i)} - \mathit{A}_{\text{vaso},t}^{(i)}\bigr|,
\end{equation}
where \(\mathit{\hat{A}}\) denotes the actions recommended by the RL policy and \(\mathit{A}\) denotes the actions taken by the clinician. 
A weighted average of \(m_{\text{iv}}^{(i)}\) and \(m_{\text{vaso}}^{(i)}\) is used to compute the overall discrepancy:

\begin{equation}
m^{(i)} 
= \alpha \, m_{\text{iv}}^{(i)} + \beta \, m_{\text{vaso}}^{(i)}, 
\quad
\alpha + \beta = 1,\;\alpha, \beta \ge 0.
\end{equation} 

To evaluate the clinical impact of adherence to the RL policy, we compare survival rates between episodes with low and high policy discrepancies. 
We identify the most extreme \(N\%\) of episodes by sorting the discrepancy scores \(m^{(i)}\) and computing the \(p = N/2\) and \((100 - p)\) percentiles, denoted \(Q_p\) and \(Q_{100-p}\), where \(p\) is a hyperparameter. 
The low and high-discrepancy cohort is defined as:
\begin{equation}
    \begin{aligned}
L = \bigl\{\,i \mid m^{(i)} \le Q_{p}\bigr\}, &&  H = \bigl\{\,i \mid m^{(i)} \ge Q_{100-p}\bigr\}.
    \end{aligned}
\end{equation}

These cohorts represent the lower and upper extremes of the discrepancy distribution.
Survival status for each episode is encoded as a binary indicator \(S^{(i)}\in\{0,1\}\), where 0 indicates mortality. 
Survival rate is defined as follows:

\begin{equation}
\begin{aligned}
\mathrm{High\text{-}BDESR} 
= \frac{1}{|H|} \sum_{i \in H} \mathbf{1}\{S^{(i)}=1\}, \\
\quad
\mathrm{Low\text{-}BDESR} 
= \frac{1}{|L|} \sum_{i \in L} \mathbf{1}\{S^{(i)}=1\}.
\end{aligned}
\end{equation}
BDESR facilitates a more precise evaluation of the clinical effectiveness of an RL policy by enabling a comparison of survival between patients with well-behaved (Low-BDESR) and significantly deviant (High-BDESR) behaviors.

\subsection{Implementation details}
All experiments were conducted with fixed hyperparameter settings for consistency. 
We trained all models using a batch size of 256, a learning rate of 0.0001, and the Adam optimizer.
The BCQ threshold was set to 0.3, and the CQL regularization coefficient to 2.0. 
Both BCQ and CQL employed a Dueling DQN architecture \cite{ddqn}.
The structured data-only RL model employed a neural network consisting of three fully connected layers with 512 units each. 
The text-based model employed an LLM for embedding, integrated with a Q-network.
Textual embeddings were obtained from the LLM with all weights frozen during training.

\section{Results and Discussion}

\begin{table*}[ht]
\centering
\setlength{\tabcolsep}{4pt}
\resizebox{\textwidth}{!}{%
\begin{tabular}{llcccccccc}
\toprule
 &  & \multicolumn{2}{c}{\textbf{Structured data}}
      & \multicolumn{3}{c}{\textbf{Text}}
      & \multicolumn{3}{c}{\textbf{Multimodal (MORE-CLEAR)}} \\
\cmidrule(lr){3-4} \cmidrule(lr){5-7} \cmidrule(lr){8-10}
\textbf{Dataset} & \textbf{Metric}
  & \textbf{BCQ} & \textbf{CQL}
  & \textbf{Bert} & \textbf{CB} & \textbf{Llama}
  & \textbf{Bert+CQL} & \textbf{CB+CQL} & \textbf{Llama+CQL} \\
\midrule
\multirow{4}{*}{MIMIC-III}
 &  $\uparrow$ OPERA & $0.889\pm0.02$ & $2.925\pm0.03$
         & $0.512\pm0.09$ & $0.607\pm0.09$ & $0.450\pm0.09$
         & $\underline{3.335\pm0.14}$ & $3.329\pm0.10$ & $\underline{3.382\pm0.14}$ \\
 &  $\uparrow$ DR    & $0.917\pm0.04$ & $2.419\pm0.05$
         & $0.574\pm0.08$ & $0.587\pm0.07$ & $0.546\pm0.10$
         & $2.968\pm0.10$ & $\underline{2.975\pm0.06}$ & $\underline{3.007\pm0.09}$ \\
 &  $\uparrow$ FQE   & $0.472\pm0.12$ & $0.622\pm0.07$
         & $0.320\pm0.28$ & $0.485\pm0.32$ & $0.181\pm0.08$
         & $\underline{1.382\pm0.74}$ & $\underline{1.346\pm0.38}$ & $0.899\pm0.22$ \\
 &  $\uparrow$ WIS   & $0.563\pm0.25$ & $0.682\pm0.03$
         & $0.721\pm0.08$ & $\underline{0.738\pm0.02}$ & $\underline{0.728\pm0.08}$
         & $0.693\pm0.02$ & $0.702\pm0.04$ & $0.683\pm0.03$ \\
\midrule
\multirow{4}{*}{MIMIC-IV}
 &  $\uparrow$ OPERA & $0.885\pm0.09$ & $\underline{3.862\pm0.04}$
         & $0.436\pm0.07$ & $0.461\pm0.07$ & $0.714\pm0.36$
         & $3.861\pm0.16$ & $\underline{3.877\pm0.16}$ & $3.810\pm0.14$ \\
 &  $\uparrow$ DR    & $0.963\pm0.05$ & $\underline{3.791\pm0.03}$
         & $0.521\pm0.24$ & $0.866\pm0.18$ & $0.750\pm0.29$
         & $3.648\pm0.10$ & $\underline{3.677\pm0.09}$ & $3.596\pm0.12$ \\
 &  $\uparrow$ FQE   & $0.466\pm0.10$ & $1.243\pm0.13$
         & $0.406\pm0.22$ & $0.210\pm0.05$ & $0.535\pm0.24$
         & $\underline{2.597\pm2.07}$ & $\underline{7.522\pm3.33}$ & $1.322\pm0.35$ \\
 &  $\uparrow$ WIS   & $0.731\pm0.08$ & $0.753\pm0.03$
         & $0.566\pm0.31$ & $0.736\pm0.12$ & $0.739\pm0.12$
         & $0.734\pm0.04$ & $\underline{0.758\pm0.04}$ & $\underline{0.766\pm0.04}$ \\
\midrule
\multirow{4}{*}{Private Dataset}
 &  $\uparrow$ OPERA & $0.782\pm0.05$ & $2.390\pm0.22$
         & $0.602\pm0.10$ & $0.599\pm0.12$ & $0.717\pm0.22$
         & $\underline{2.781\pm0.17}$ & $\underline{2.930\pm0.14}$ & $1.969\pm0.28$ \\
 &  $\uparrow$ DR    & $0.952\pm0.07$ & $2.487\pm0.20$
         & $0.752\pm0.09$ & $0.601\pm0.09$ & $0.689\pm0.06$
         & $\underline{2.790\pm0.23}$ & $\underline{2.739\pm0.09}$ & $1.893\pm0.27$ \\
 &  $\uparrow$ FQE   & $0.591\pm0.13$ & $1.290\pm0.28$
         & $0.692\pm0.15$ & $0.510\pm0.33$ & $0.585\pm0.07$
         & $1.517\pm0.71$ & $\underline{1.681\pm0.96}$ & $\underline{1.811\pm0.42}$ \\
 &  $\uparrow$ WIS   & $0.675\pm0.26$ & $0.685\pm0.02$
         & $0.422\pm0.22$ & $0.413\pm0.28$ & $0.464\pm0.40$
         & $\underline{0.742\pm0.01}$ & $\underline{0.729\pm0.04}$ & $0.627\pm0.25$ \\
\bottomrule
\end{tabular}%
}
\caption{OPE metric performance across all modalities. The means and standard deviations of the results across five seeds are reported. The best two results are underlined for each metric. CB: Clinical Bert, Llama: Llama3.1-8B}
\label{tab_ope}
\end{table*}

\begin{table*}[ht]
\centering
\setlength{\tabcolsep}{4pt}
\resizebox{\textwidth}{!}{%
\begin{tabular}{llcccccccc}
\toprule
 &  & \multicolumn{2}{c}{\textbf{Structured data}}
      & \multicolumn{3}{c}{\textbf{Text}}
      & \multicolumn{3}{c}{\textbf{Multimodal (MORE-CLEAR)}} \\
\cmidrule(lr){3-4} \cmidrule(lr){5-7} \cmidrule(lr){8-10}
\textbf{Dataset} & \textbf{Metric}
  & \textbf{BCQ} \cite{6:choi:2024}  & \textbf{CQL} 
  & \textbf{Bert} & \textbf{CB} & \textbf{Llama}
  & \textbf{Bert+CQL} & \textbf{CB+CQL} & \textbf{Llama+CQL} \\
\midrule
\multirow{2}{*}{MIMIC-III}
 & Low-BDESR
   & $0.848\pm0.04$ & $0.847\pm0.03$
   & $0.834\pm0.03$ & $0.815\pm0.03$ & $0.836\pm0.01$
   & $0.832\pm0.02$ & $\underline{0.861\pm0.03}$ & $\underline{0.862\pm0.02}$ \\
 & High-BDESR 
   & $0.744\pm0.06$ & $0.633\pm0.02$
   & $0.682\pm0.06$ & $0.660\pm0.06$ & $0.722\pm0.08$
   & $0.631\pm0.01$ & $\underline{0.619\pm0.02}$ & $\underline{0.614\pm0.01}$ \\
\midrule
\multirow{2}{*}{MIMIC-IV}
 & Low-BDESR 
   & $0.843\pm0.05$ & $\underline{0.851\pm0.01}$
   & $0.827\pm0.02$ & $0.839\pm0.02$ & $0.790\pm0.03$
   & $0.840\pm0.01$ & $\underline{0.856\pm0.02}$ & $0.837\pm0.02$ \\
 & High-BDESR 
   & $0.733\pm0.04$ & $0.635\pm0.02$
   & $0.688\pm0.02$ & $0.700\pm0.02$ & $0.560\pm0.02$
   & $0.631\pm0.03$ & $\underline{0.618\pm0.03}$ & $\underline{0.569\pm0.01}$ \\
\midrule
\multirow{2}{*}{Private Dataset}
 & Low-BDESR 
   & $0.889\pm0.08$ & $0.893\pm0.06$
   & $0.856\pm0.06$ & $0.901\pm0.00$ & $\underline{0.930\pm0.05}$
   & $0.810\pm0.09$ & $\underline{0.921\pm0.12}$ & $0.891\pm0.04$ \\
 & High-BDESR 
   & $0.722\pm0.13$ & $\underline{0.606\pm0.05}$
   & $0.636\pm0.12$ & $\underline{0.578\pm0.14}$ & $0.644\pm0.09$
   & $0.644\pm0.09$ & $0.614\pm0.06$ & $0.711\pm0.06$ \\
\bottomrule
\end{tabular}%
}
\caption{Performance of the BDESR metric across all modalities. The means and standard deviations of the results across five seeds are reported. The top two results are underlined for each metric.}
\label{tab_bdesr}
\end{table*}

\subsection{Performance Comparison Across Modalities}
The OPE and BDESR performances for each modality are reported in Tables \ref{tab_ope} and \ref{tab_bdesr}, respectively. 
Table \ref{tab_ope} shows that the MORE-CLEAR generally outperforms unimodal baselines across a broad evaluation metrics. 
The multimodal models (Bert+CQL, CB+CQL, and Llama+CQL) outperform both structured data-only and text-only baselines, particularly concerning the OPERA (0.450 \(\to\) 3.382 in Llama+CQL), DR (0.546 \(\to\) 3.007 in Llama+CQL), and FQE (0.320 \(\to\) 1.382).
Dataset-specific trends further reinforce the advantage of the MORE-CLEAR framework. 
On MIMIC-III, Llama+CQL and Bert+CQL show the highest scores on OPERA (3.382), DR (3.007), and FQE (1.382), respectively.
Furthermore, Llama+CQL achieves the best outcomes for WIS (0.766), whereas CB+CQL dominates the outstanding performance on the FQE score of 7.522 in MIMIC-IV.
On the private dataset, the superiority of multimodal models remains evident.
Across all settings, CB+CQL emerges as the most robust combination, with Bert+CQL also exhibiting strong and consistent results. 
Combining text representations with conventional structured data-based RL algorithms provides clear performance benefits.
These results also provide substantial evidence supporting the efficacy of the MORE-CLEAR framework, which offers a meaningful synergy of multimodal data.

Table \ref{tab_bdesr} presents estimated survival rates based on the BDESR, which stratifies patient episodes into two cohorts: those closely aligned with the RL policy’s recommended actions (Low-BDESR) and those that deviate significantly (High-BDESR). 
We set \(p\) to 20.
Across all datasets, survival rates for the Low-BDESR group were consistently higher than those for High-BDESR, indicating that the RL policy has been trained in a clinically beneficial direction. 
Within the Low-BDESR group, multimodal configurations demonstrated superior survival rates in every dataset, with CB+CQL consistently achieving the highest rates of 0.861, 0.856, and 0.921 for MIMIC-III, MIMIC-IV, and private dataset, respectively. 
This finding suggests that the integration of clinical notes and structured data contributes meaningfully to the identification of effective treatment strategies.
Furthermore, structured data-based models outperformed their text-only counterparts, suggesting that structured data may provide more informative patient state representations for policy learning than the clinical note itself. 
Conversely, survival rates in the High-BDESR group exhibited a decline across all modalities, likely attributable to including a greater number of trajectories that culminated in mortality. 
The evidence suggests that the RL policy is inclined to suggest courses of action that diverge from adverse outcomes, thereby indicating its capacity to circumvent detrimental treatment patterns.

\subsection{Contextual Representation Performance}
We evaluate four strategies for integrating clinical notes: (1) raw note embedding (raw), (2) simple token imputation (impute), (3) concatenation within a fixed time window (stack), and (4) encoding the initial note as a static context vector (context). 
In raw note embedding, the note is utilized as input if it exists within the time frame (4 hours); otherwise, no textual input is provided.
Simple token imputation involves forward-filling the note if it does not exist in the current time frame.
Concatenation within a fixed time window stacks notes of the previous 12 hours (window size(\(W\))=3; see Appendix B for more details).
Figure \ref{note_integration} shows experimental results across several OPE metrics from CB+CQL policy learning.
OPERA scores consistently increase from raw to context in MIMIC-III ($\approx$2.6\(\to\)$\approx$3.3) and MIMIC-IV($\approx$3.3\(\to\)$\approx$3.8). 
On the PD dataset, context achieves the highest OPERA score ($\approx$2.7). 
Context also yields the best performance on the DR metric, particularly in PD.
FQE differences are negligible in MIMIC-III but favor context in MIMIC-IV, albeit with high variance. 
In PD, the stack method attains the highest FQE score.
WIS scores remain comparable across all approaches, with context performing slightly better. 
These results indicate that leveraging the initial clinical note as a context vector using our context-aware gated fusion improves policy performance, though some metrics remain sensitive to high variance.

\begin{figure*}[htbp]
  \centering
  \makebox[\textwidth][c]{%
    \begin{subfigure}[b]{0.23\textwidth}
      \centering
      \includegraphics[width=\linewidth]{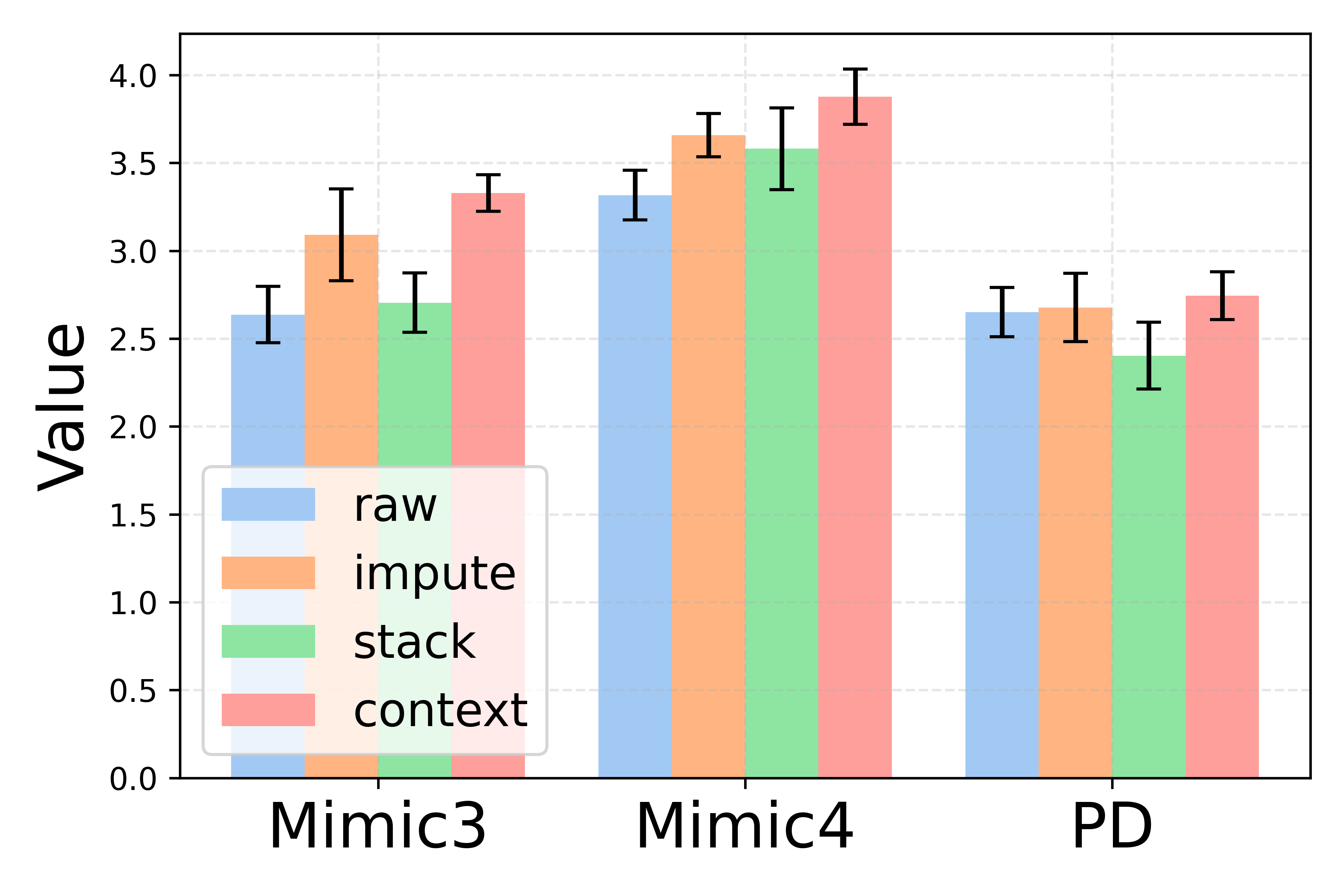}
      \caption{OPERA}
      \label{fig1-opera}
    \end{subfigure}\hspace{0.01\textwidth}%
    \begin{subfigure}[b]{0.23\textwidth}
      \centering
      \includegraphics[width=\linewidth]{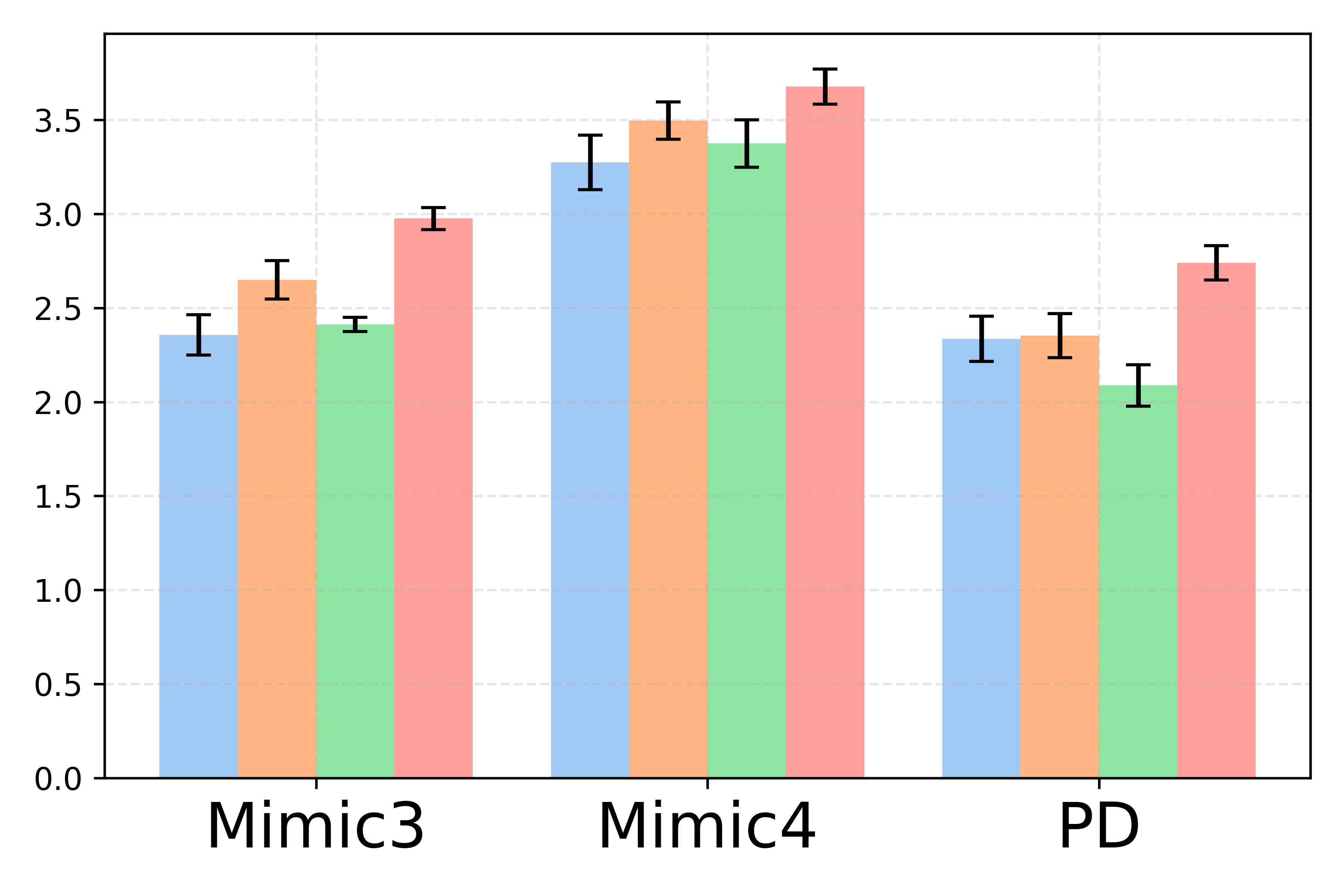}
      \caption{DR}
      \label{fig1-dr}
    \end{subfigure}\hspace{0.01\textwidth}%
    \begin{subfigure}[b]{0.23\textwidth}
      \centering
      \includegraphics[width=\linewidth]{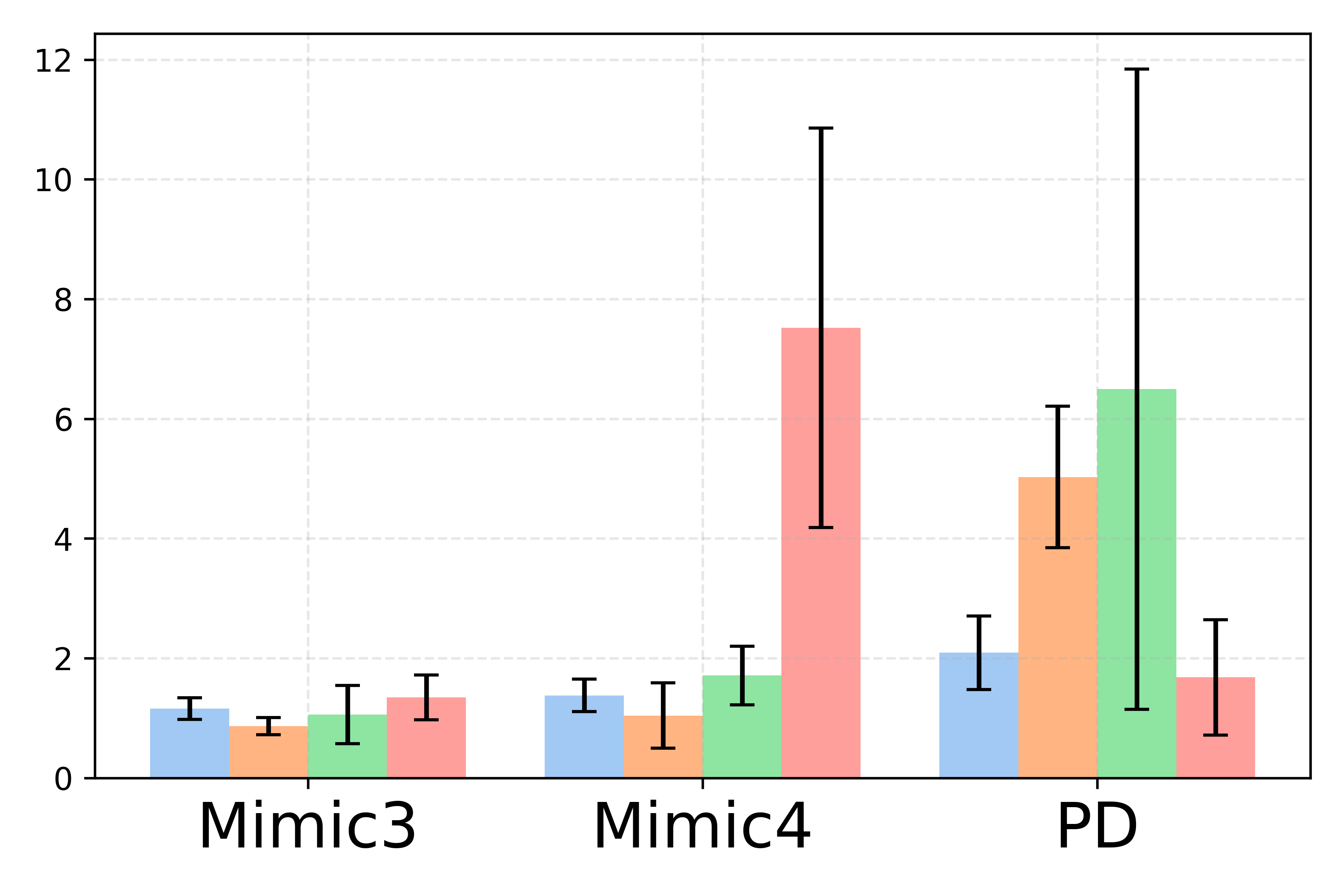}
      \caption{FQE}
      \label{fig1-fqe}
    \end{subfigure}\hspace{0.01\textwidth}%
    \begin{subfigure}[b]{0.23\textwidth}
      \centering
      \includegraphics[width=\linewidth]{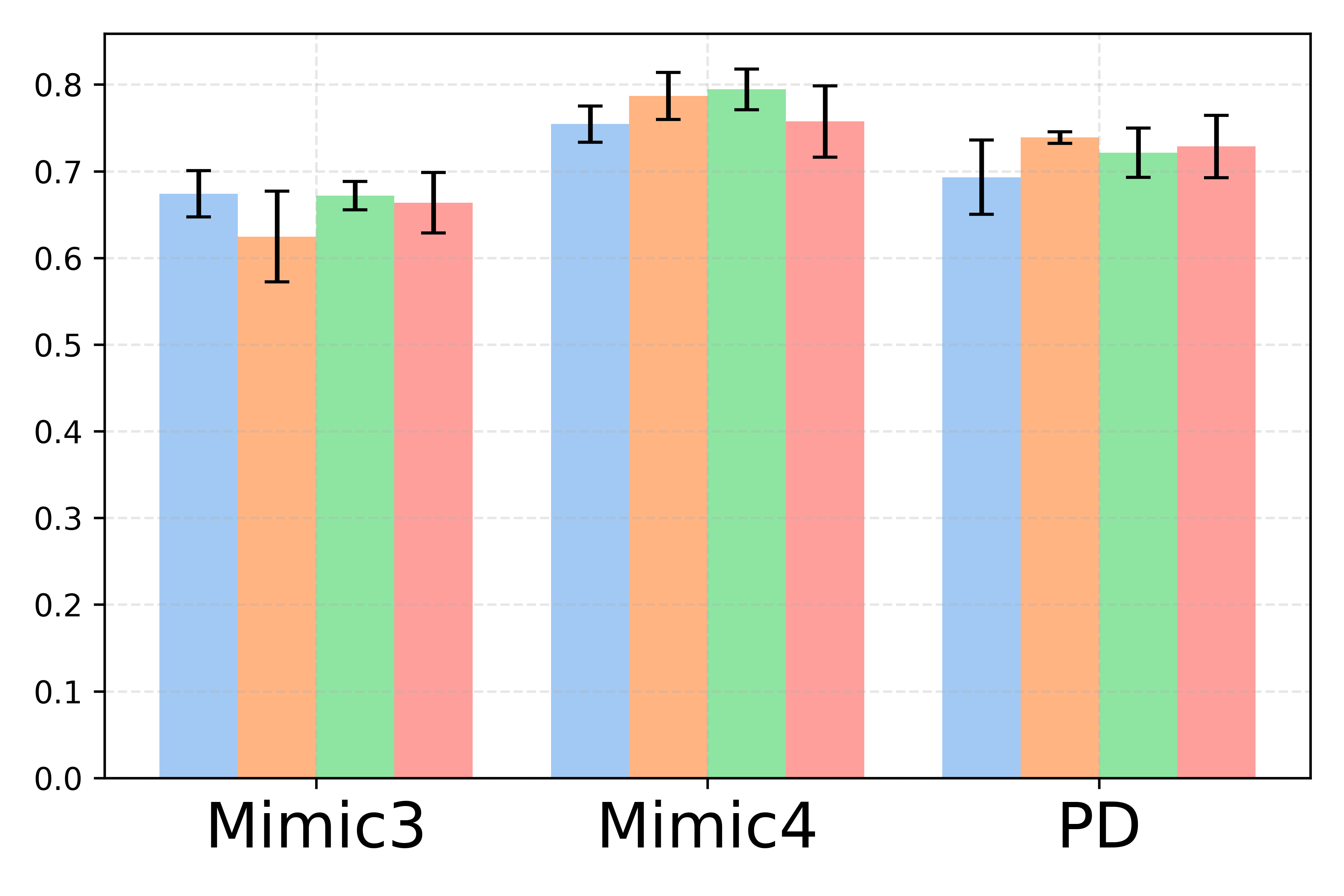}
      \caption{WIS}
      \label{fig1-wis}
    \end{subfigure}%
  }
  \caption{Performance of policies under different clinical note integration strategies}
  \label{note_integration}
\end{figure*}

\subsection{External Validation}
Table \ref{tab:cross validation results} reports cross-dataset validation results of the MORE-CLEAR framework in conjunction with CB+CQL.
When evaluated on MIMIC-IV, the policy trained on MIMIC-III yields OPERA 3.004, FQE 2.116, and WIS 0.689, while the policy trained using PD achieves a slightly higher OPERA, a substantially lower FQE, and a comparable WIS.
In the reverse setting, the MIMIC-III-based policy demonstrates suboptimal performance on PD, while the MIMIC-IV-based policy exhibits marginally superior generalization.
On MIMIC-III, the policy trained with PD yields OPERA 2.830, FQE 1.150, and WIS 0.601, while the policy trained with MIMIC-IV attains the highest OPERA score of 3.974 across all settings.
Overall, the policy trained using MIMIC-IV demonstrates consistent performance in terms of high expected returns and low-variance estimates across datasets, thereby indicating superior generalization. 
Conversely, the MIMIC-III-based policy demonstrates a high degree of sensitivity to distributional shifts, resulting in performance degradation in the external dataset.

\begin{figure}[htbp]
  \centering
  \begin{subfigure}[b]{0.48\columnwidth}
    \centering
    \includegraphics[width=\linewidth]{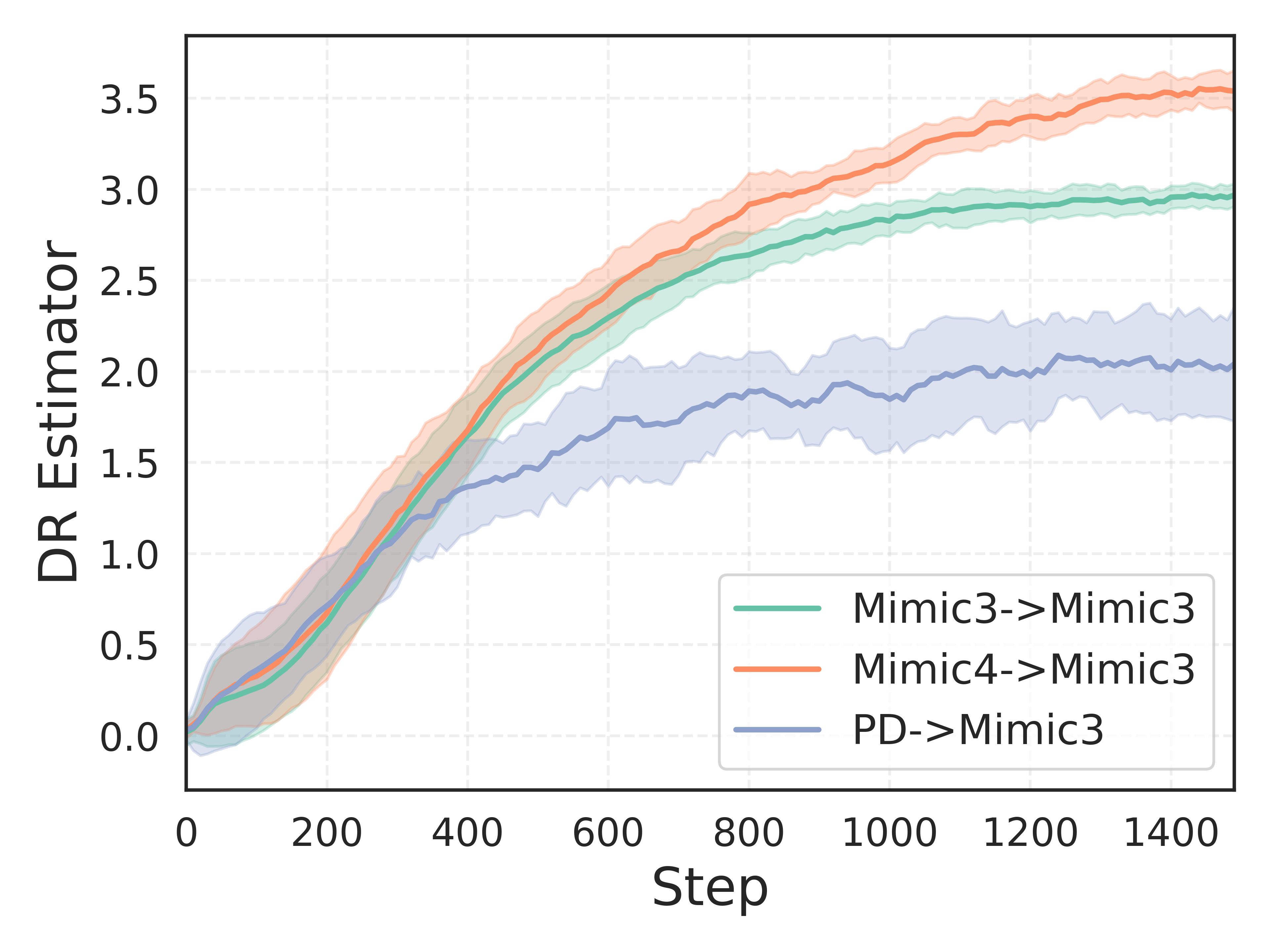}
    \caption{MIMIC-III}
    \label{fig:sub1}
  \end{subfigure}\hfill
  \begin{subfigure}[b]{0.48\columnwidth}
    \centering
    \includegraphics[width=\linewidth]{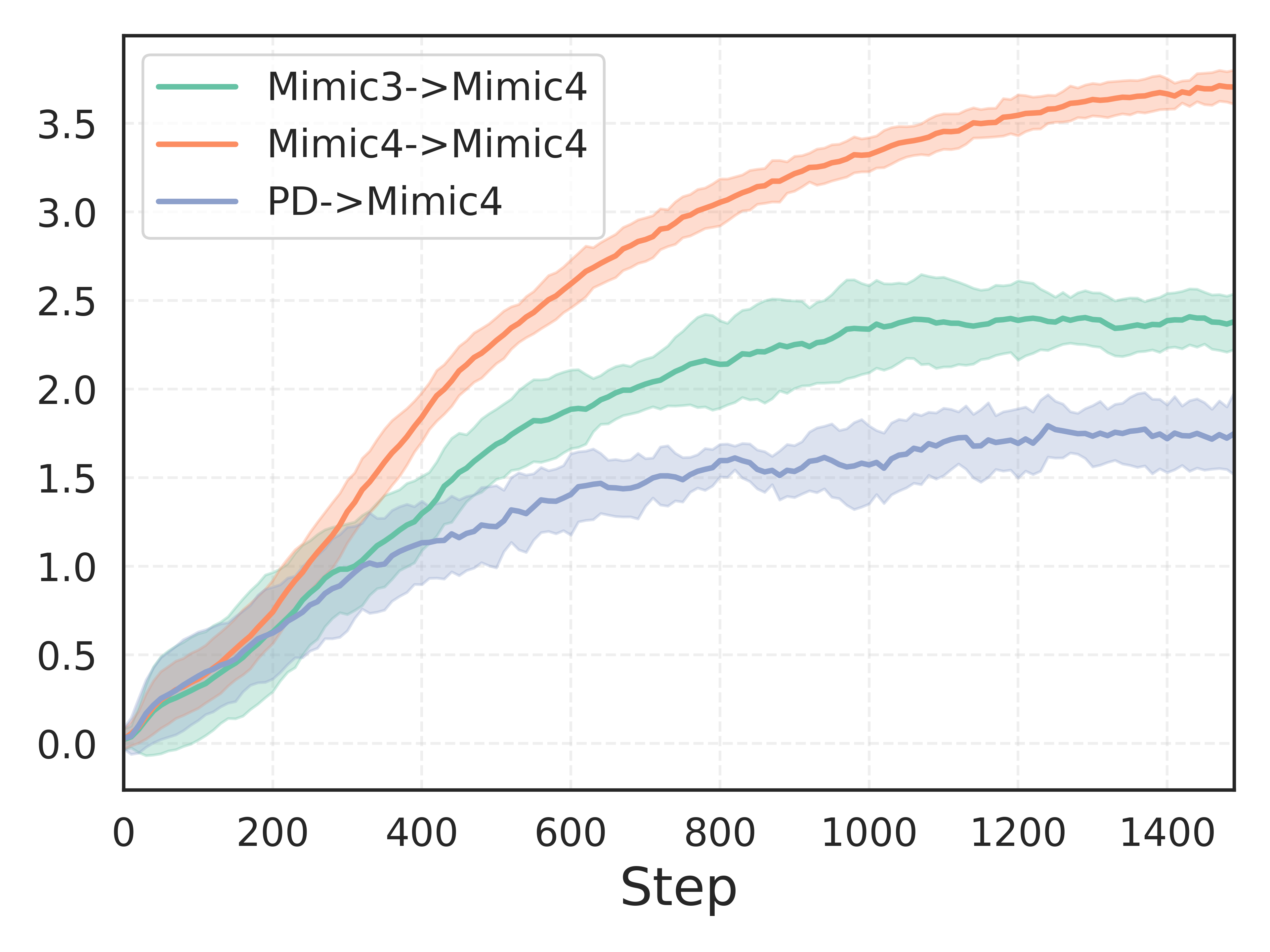}
    \caption{MIMIC-IV}
    \label{fig:sub2}
  \end{subfigure}
  \begin{subfigure}[b]{0.48\columnwidth}
    \centering
    \includegraphics[width=\linewidth]{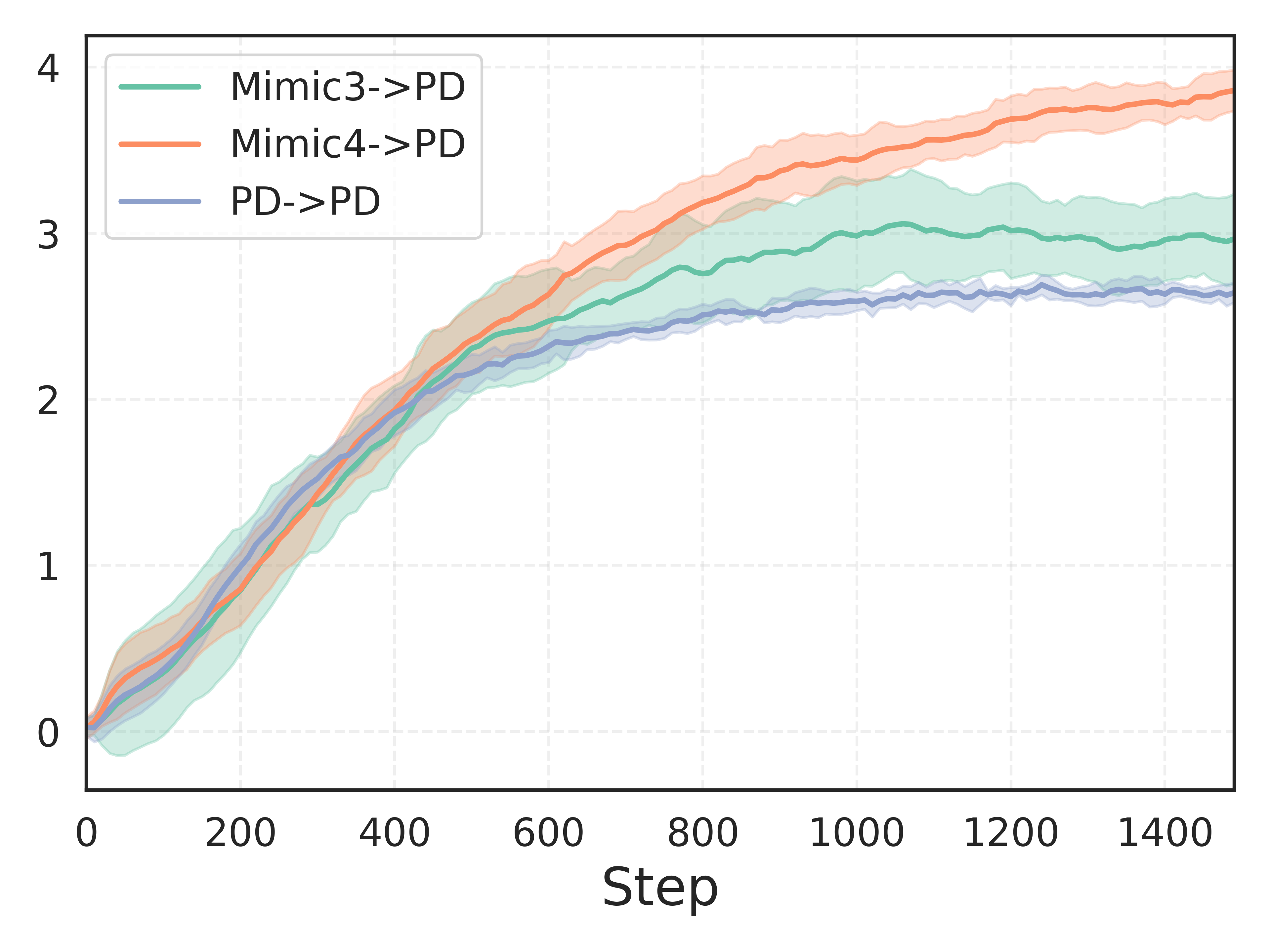}
    \caption{PD}
    \label{fig:sub2}
  \end{subfigure}
  \caption{DR estimator performance across training iterations under cross-dataset validation}
  \label{fig:cross-dr}
\end{figure}

In order to assess whether policies improve consistently across different datasets, the DR estimates are further visualized across training iterations in Figure \ref{fig:cross-dr}.
Our observations indicate that policy learning exhibits a positive reinforcement trend across all datasets.
It is noteworthy that the model trained on MIMIC-IV consistently attains the highest performance when evaluated on other datasets, followed by models trained on MIMIC-III and PD, respectively. 
This finding indicates that the data distribution of MIMIC-IV effectively captures the underlying characteristics of both MIMIC-III and PD. 

The results of these trends can be interpreted in light of the characteristics of each dataset.
For instance, the data collection process for MIMIC-IV has been more refined in comparison to that of MIMIC-III, resulting in higher-quality data \cite{41:johnson:2023}. 
The clinical notes themselves exhibit greater consistency, as MIMIC-IV is collected on a radiology report basis.
Furthermore, the MIMIC-IV, which was collected from 2008 to 2022, may contain a more homogeneous sepsis treatment trajectory than the MIMIC-III, which was collected from 2001 to 2012, because it has more data since the treatment guidelines were revised to emphasize the importance of sepsis \cite{fluid_volume_1}.

As such, offline RL is susceptible to distributional shifts, influenced by the nature of the data.
Rigorous validation in diverse cohorts is imperative prior to its implementation in a clinical setting.
We have demonstrated that our proposed MORE-CLEAR framework performs robustly in cross-validation based on a wide variety of data sets, including those from disparate collection periods and countries, thereby substantiating the model's viability for implementation in real-world clinical settings.

\begin{table}[ht]
  \centering
  \captionsetup{skip=10pt}
  \scriptsize   
  \begin{tabular}{lccc}
    \toprule
    \textbf{Train$\rightarrow$Test} & \textbf{OPERA} & \textbf{FQE} & \textbf{WIS} \\
    \midrule
    MIMIC-III $\rightarrow$ MIMIC-IV  & 3.004 ± 0.148 & 2.116 ± 1.293 & 0.689 ± 0.033 \\
    PD   $\rightarrow$ MIMIC-IV  & 3.107 ± 0.303 & 1.188 ± 0.110 & 0.677 ± 0.017 \\
    \midrule
    MIMIC-III $\rightarrow$ PD     & 2.011 ± 0.162 & 0.835 ± 0.363 & 0.731 ± 0.065 \\
    MIMIC-IV $\rightarrow$ PD    & 2.909 ± 0.148 & 0.897 ± 0.395 & 0.744 ± 0.049 \\
    \midrule
    PD   $\rightarrow$ MIMIC-III  & 2.830 ± 0.088 & 1.150 ± 0.223 & 0.601 ± 0.038 \\
    MIMIC-IV $\rightarrow$ MIMIC-III  & 3.974 ± 0.115 & 1.488 ± 0.576 & 0.649 ± 0.018 \\
    \bottomrule
  \end{tabular}
\caption{Cross‐dataset validation results (CB+CQL)}
\label{tab:cross validation results} 
\end{table}

\subsection{Overestimation Analysis}
\begin{figure}[htbp]
  \centering
  \begin{subfigure}[b]{0.48\columnwidth}
    \centering
    \includegraphics[width=\linewidth]{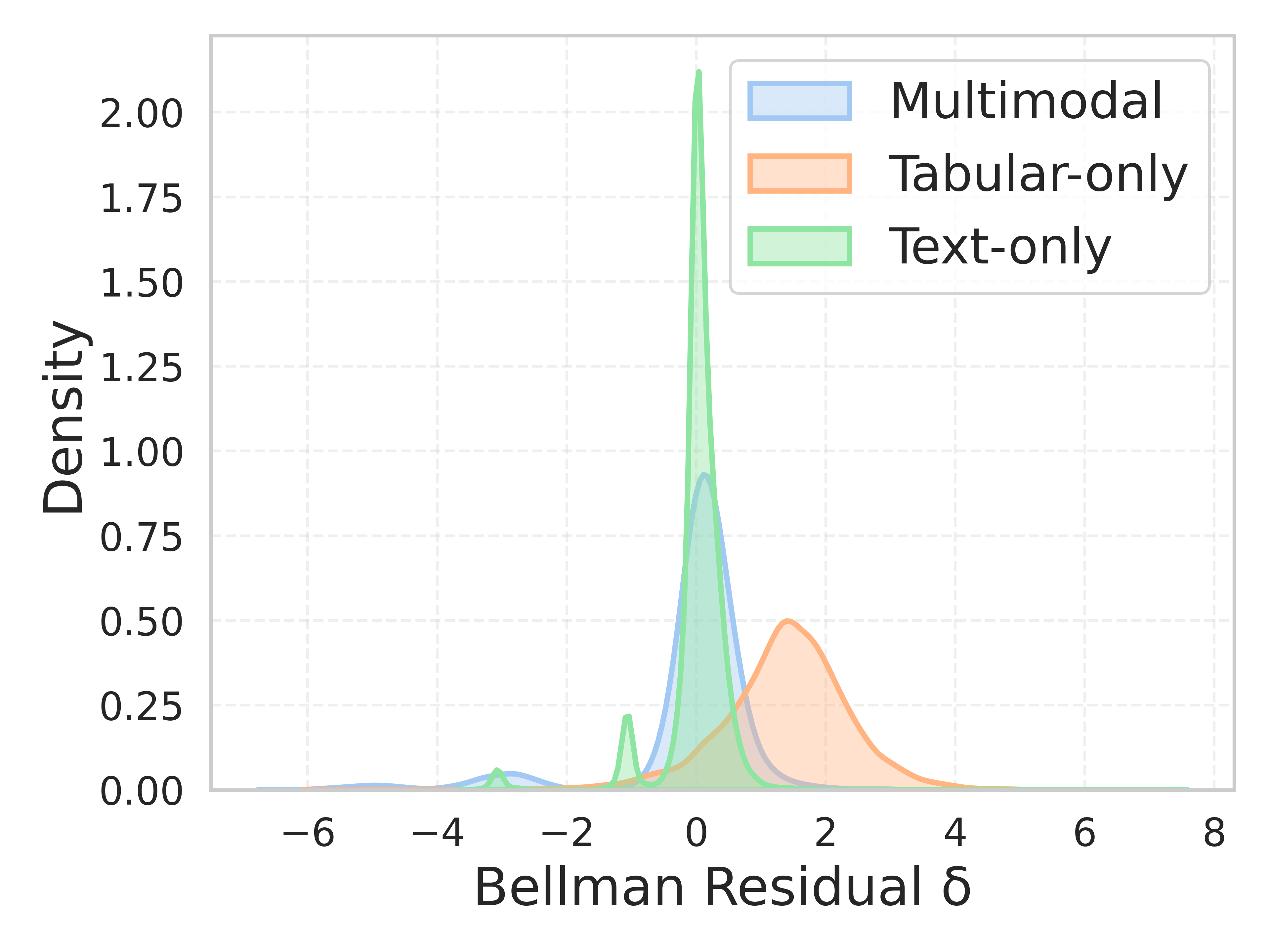}
    \caption{MIMIC-III}
    \label{fig:overestmation:a}
  \end{subfigure}\hfill
  \begin{subfigure}[b]{0.48\columnwidth}
    \centering
    \includegraphics[width=\linewidth]{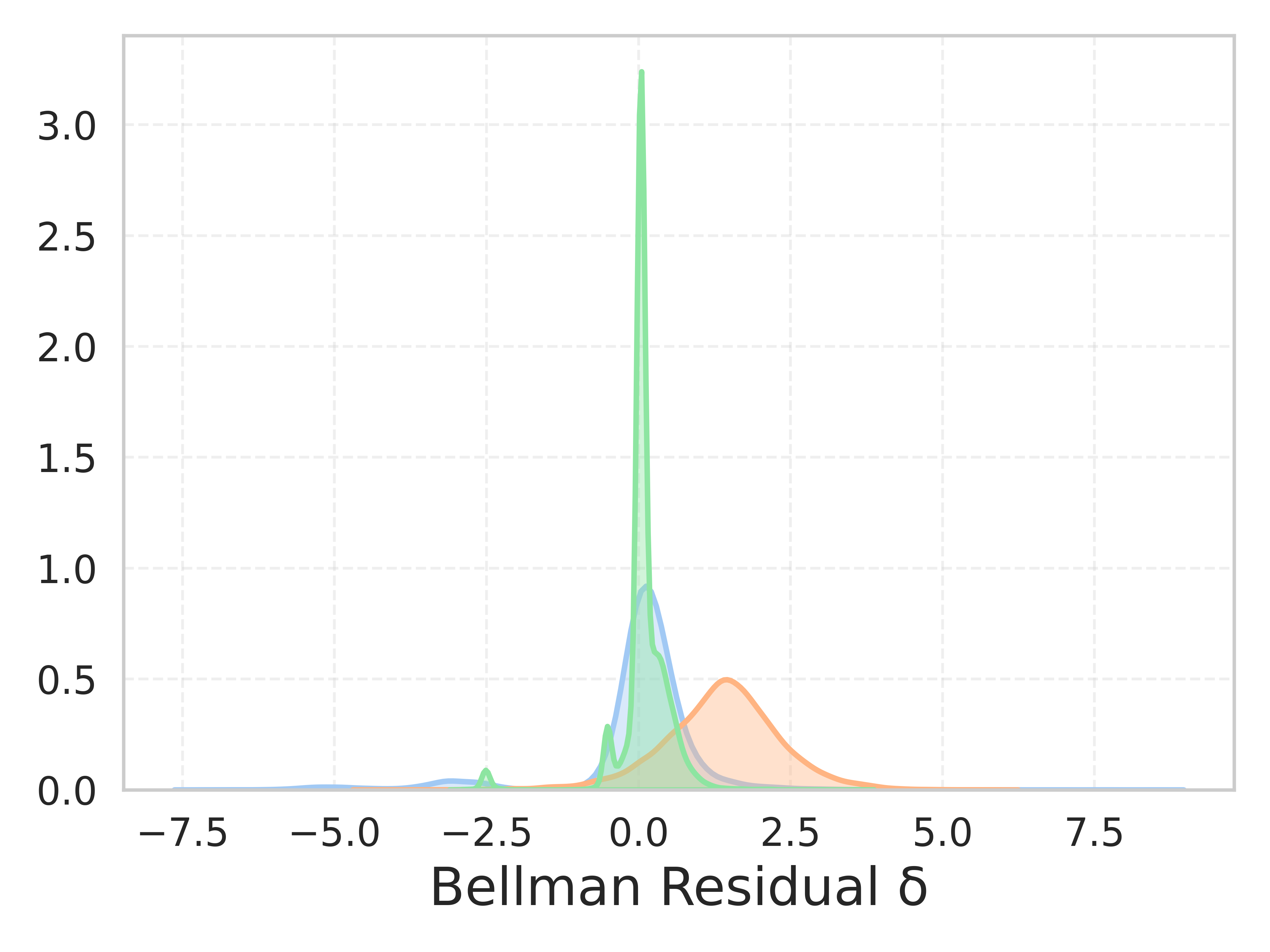}
    \caption{MIMIC-IV}
    \label{fig:overestmation:b}
  \end{subfigure}
  \begin{subfigure}[b]{0.48\columnwidth}
    \centering
    \includegraphics[width=\linewidth]{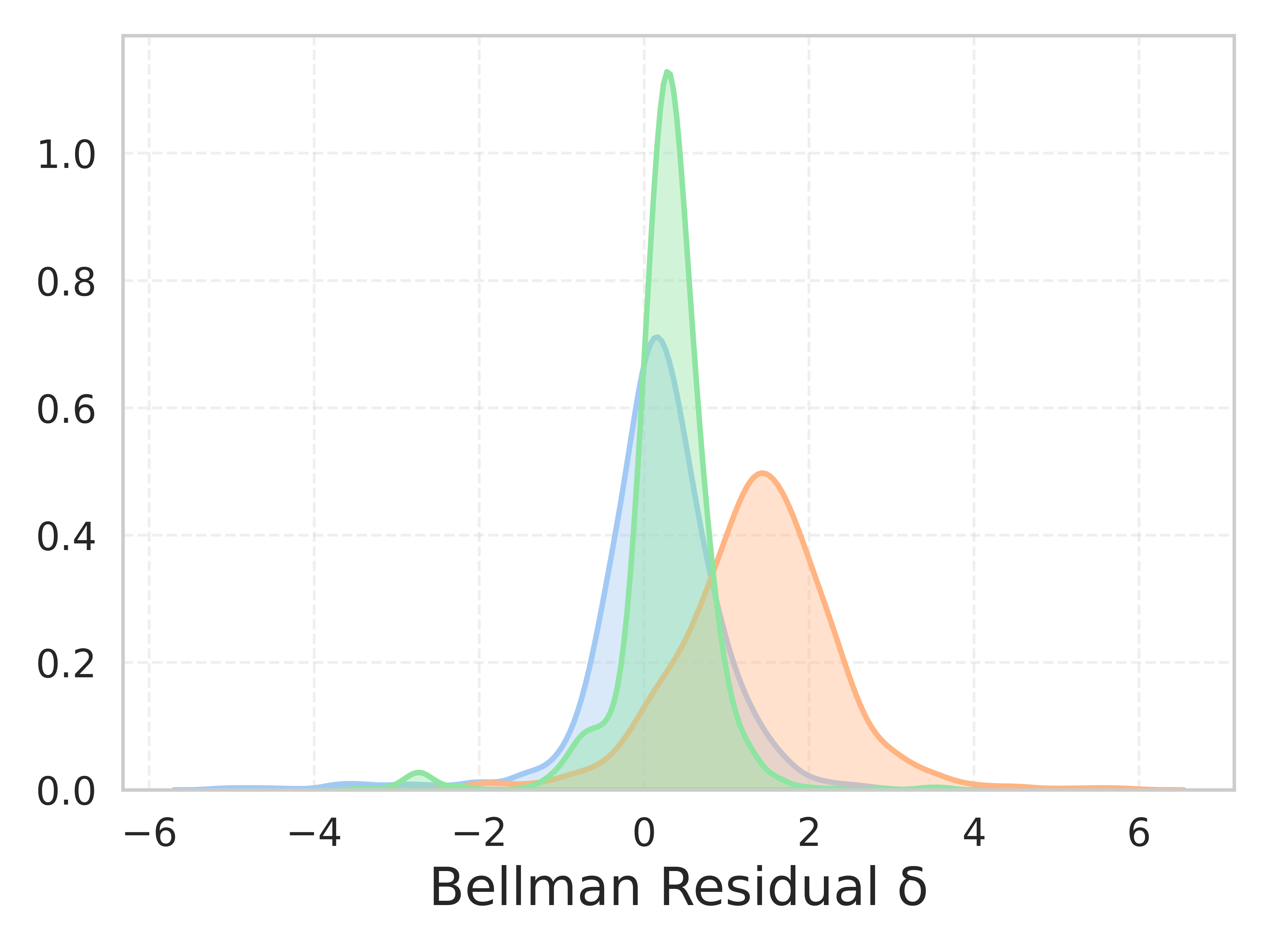}
    \caption{PD}
    \label{fig:overestmation:c}
  \end{subfigure}
  \caption{Bellman residual distributions across modalities}
  \label{fig:overestimation}
\end{figure}

Figure \ref{fig:overestimation} presents the estimated probability density functions of the Bellman residuals for three policies: Multimodal (blue), Structured data-only (orange), and Text-only (green). 
It directly compares overestimation bias and estimation stability across datasets.
In Figure \ref{fig:overestmation:a}, the multimodal policy exhibits a distribution tightly concentrated around zero with minimal variance, indicating low bias and high reliability in value estimation. 
By contrast, the density of structured data-only policies is shifted to the right and substantially wider, reflecting a pronounced positive bias (overestimation) and unstable predictions.
The text-only policy shows a sharp peak at zero but with a narrow spread, suggesting overconfident estimates that may under-represent actual estimation error.
Figure \ref{fig:overestmation:b} and Figure \ref{fig:overestmation:c} further demonstrate that the structured data-only policy exhibits overestimation, while the text-only policy shows excessively high confidence in its estimates. 
In contrast, the multimodal approach achieves a more balanced value estimation between exploration and exploitation.
These results demonstrate that integrating structured data with clinical notes via the MORE-CLEAR framework substantially reduces both bias and variance in Bellman residuals, yielding more stable and trustworthy value‐function estimates than either unimodal strategy.

\subsection{Summarization Performance}
\begin{figure}[htb!]
  \centering
  \begin{subfigure}[b]{0.48\columnwidth}
    \centering
    \includegraphics[width=\linewidth]{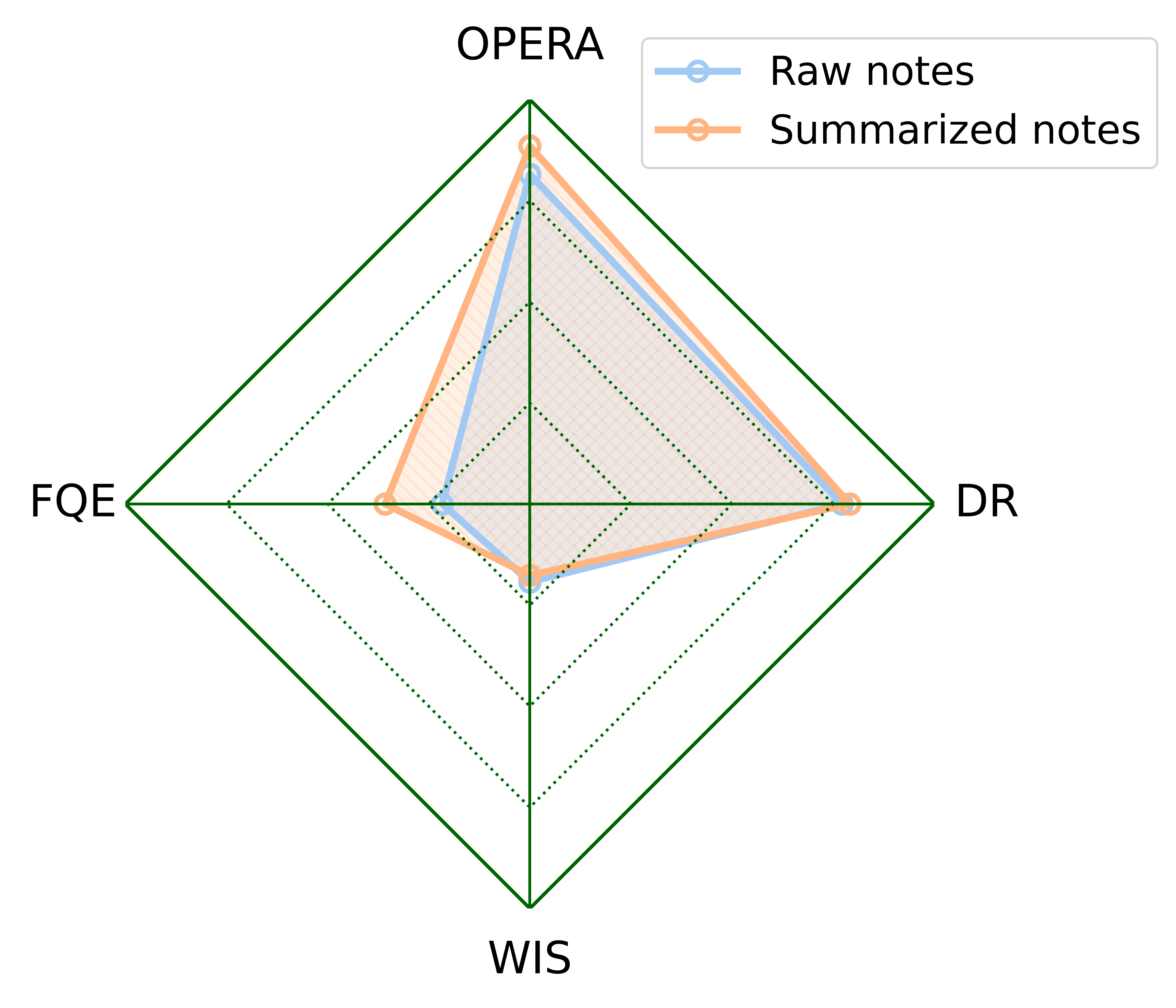}
    \caption{MIMIC-III}
    \label{fig:summarization:a}
  \end{subfigure}\hfill
  \begin{subfigure}[b]{0.48\columnwidth}
    \centering
    \includegraphics[width=\linewidth]{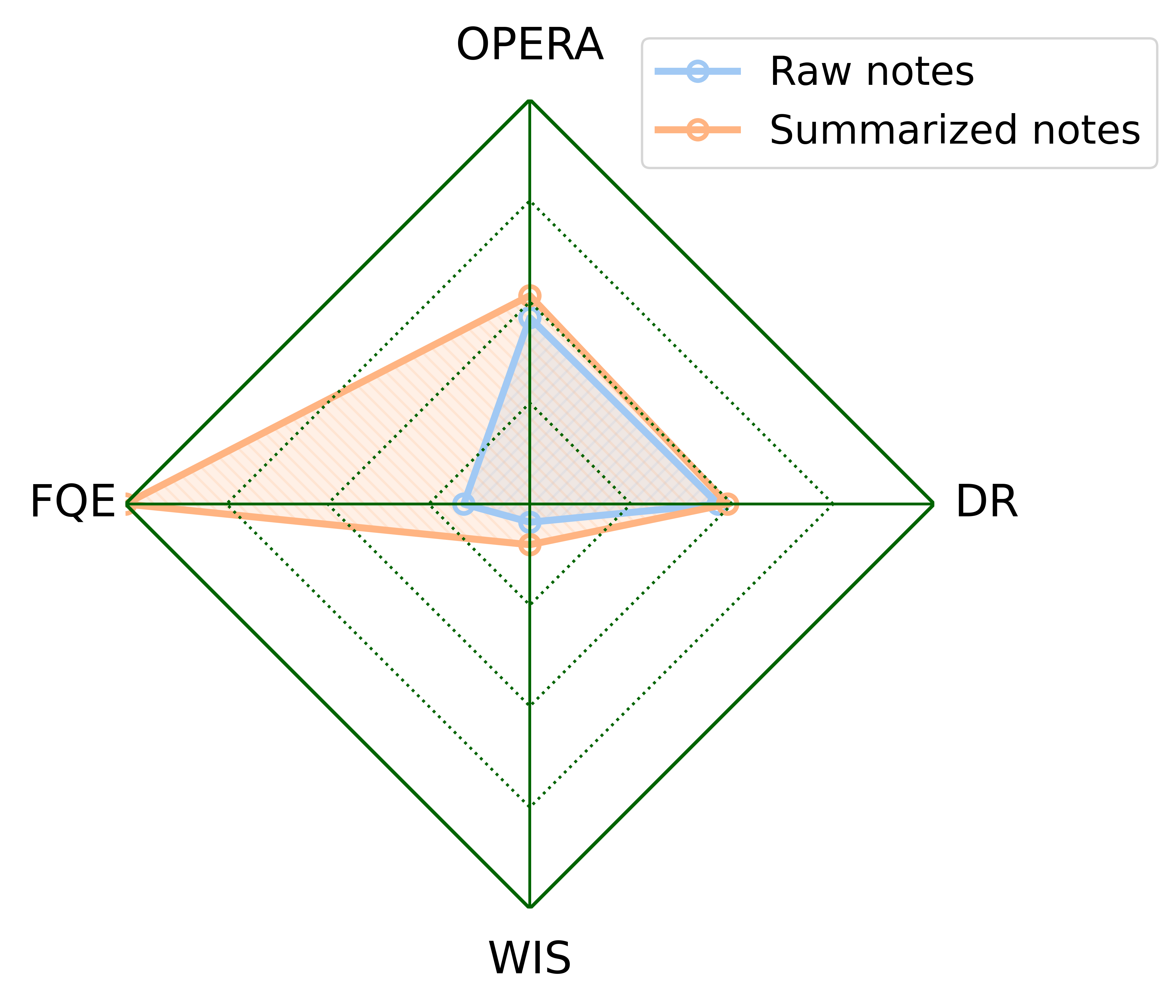}
    \caption{MIMIC-IV}
    \label{fig:summarization:b}
  \end{subfigure}
  \caption{Radar plots comparing OPE metrics between raw and summarized clinical notes}
  \label{fig:summarization}
\end{figure}

The MORE-CLEAR framework employs LLM-based summarization to mitigate the irregular distribution of unstructured information during sampling. 
Figure \ref{fig:summarization} presents results of OPE metrics for policies trained on either raw clinical notes or structured summarization notes.
The policies leveraging summarized notes exhibit consistently larger envelopes in the radar plots.
The OPERA scores extend further toward the outer grid, indicating a higher expected return under the learned policy.
The DR also improves, reflecting more accurate value estimation. 
Improvement in FQE and WIS further suggests that utilizing summarized notes leads to stable estimation of both bias and variance.
These results indicate that structured summarization of clinical notes enables more consistent learning than raw clinical notes.

\subsection{Ablation Study}

Table \ref{tab:ablation_components} presents the results of an ablation study that progressively integrates key components of the MORE-CLEAR framework, evaluating their impact on OPERA, DR, and WIS metrics.
The primary results already established the efficacy of CQL. 
Therefore, we adopt BCQ as the base algorithm to better isolate and quantify the contribution of each individual module. 
The baseline configuration, CB+BCQ, achieves a modest OPERA score of approximately 0.66 in the PD cohort, reflecting limited expected return. 
Introducing the bidirectional cross-modal attention (BCMA) module yields a substantial improvement, elevating OPERA.
The subsequent inclusion of the context-aware gated fusion (GF) mechanism further boosts OPERA by approximately 1.6 in the MIMIC-III cohort, although the improvement in PD is marginal.
The addition of GF slightly decreases the WIS score by about 0.2.
These findings suggest that BCMA is the primary contributor to performance gains. 
At the same time, GF offers auxiliary benefits by enhancing the quality of modality fusion, particularly within the MIMIC-III setting, albeit with dataset-dependent variability.

\begin{table}[ht]
  \centering
  \captionsetup{skip=10pt}
  \scriptsize            
  \setlength{\tabcolsep}{3pt}  
  \resizebox{\columnwidth}{!}{%
    \begin{tabular}{lcccccc}
      \toprule
       & \multicolumn{3}{c}{\textbf{PD}} 
       & \multicolumn{3}{c}{\textbf{MIMIC-III}} \\
      \cmidrule(lr){2-4} \cmidrule(lr){5-7}
      \textbf{Model}
        & \textbf{OPERA} & \textbf{DR} & \textbf{WIS}
        & \textbf{OPERA} & \textbf{DR} & \textbf{WIS} \\
      \midrule
      CB+BCQ
        & $0.663\pm0.04$  & $0.714\pm0.11$  & $0.528\pm0.06$
        & $0.712\pm0.13$  & $0.657\pm0.15$  & $0.791\pm0.05$  \\
      CB+BCQ+BCMA
        & $2.217\pm0.15$  & $2.262\pm0.15$  & $0.539\pm0.25$
        & $2.190\pm0.20$  & $2.090\pm0.19$  & $0.688\pm0.09$  \\

      CB+BCQ+BCMA+GF
        & $2.245\pm0.11$  & $2.395\pm0.15$  & $0.367\pm0.08$              
        & $3.730\pm0.20$  & $3.436\pm0.14$  & $0.692\pm0.05$  \\

      \bottomrule
    \end{tabular}%
  }
    \caption{Ablation study in the MORE-CLEAR framework}
    \label{tab:ablation_components}
\end{table}

\section{Conclusion}
In this work, we proposed MORE-CLEAR, a multimodal offline reinforcement learning framework for clinical decision-making by integrating structured tabular data with unstructured clinical notes.
The results demonstrate that the effective fusion of heterogeneous modalities enhances policy learning.
The incorporation of textual modalities through bidirectional cross-modal attention and context-aware gated fusion has been demonstrated to enhance the expressiveness of patient state representations, thereby leading to the development of more robust and generalizable treatment policies.
In particular, the use of context vectors from structured summarization of clinical notes contributes to policy performance.
While the present evaluation concentrated on sepsis cohorts, the proposed framework is inherently generalizable and can readily be applied to other medical tasks.

\bibliographystyle{plainnat} 
\bibliography{manuscript}

\begin{thebibliography}{68}
\providecommand{\natexlab}[1]{#1}
\providecommand{\url}[1]{\texttt{#1}}
\expandafter\ifx\csname urlstyle\endcsname\relax
  \providecommand{\doi}[1]{doi: #1}\else
  \providecommand{\doi}{doi: \begingroup \urlstyle{rm}\Url}\fi

\bibitem[Adler et~al.(2022)Adler, Wang, Mohr, and Choudhury]{31:adler:2022}
Daniel~A Adler, Fei Wang, David~C Mohr, and Tanzeem Choudhury.
\newblock Machine learning for passive mental health symptom prediction: Generalization across different longitudinal mobile sensing studies.
\newblock \emph{Plos one}, 17\penalty0 (4):\penalty0 e0266516, 2022.

\bibitem[Al~Olaimat et~al.(2024)Al~Olaimat, Bozdag, and Initiative]{12:olaimat:2024}
Mohammad Al~Olaimat, Serdar Bozdag, and Alzheimer’s Disease~Neuroimaging Initiative.
\newblock Ta-rnn: An attention-based time-aware recurrent neural network architecture for electronic health records.
\newblock \emph{Bioinformatics}, 40\penalty0 (Supplement\_1):\penalty0 i169--i179, 2024.

\bibitem[Chaabene et~al.(2023)Chaabene, Hassan, Boudaya, Chaari, and Bouaziz]{33:chaa:2023}
Siwar Chaabene, Brahim~Haroun Hassan, Amal Boudaya, Lotfi Chaari, and Bassem Bouaziz.
\newblock New mci detection method based on transformer and eeg data.
\newblock In \emph{2023 31st European Signal Processing Conference (EUSIPCO)}, pages 1200--1204. IEEE, 2023.

\bibitem[Choi et~al.(2024)Choi, Oh, Huh, Joo, Lee, You, Bae, Choi, and Kim]{6:choi:2024}
Yunho Choi, Songmi Oh, Jin~Won Huh, Ho-Taek Joo, Hosu Lee, Wonsang You, Cheng-mok Bae, Jae-Hun Choi, and Kyung-Joong Kim.
\newblock Deep reinforcement learning extracts the optimal sepsis treatment policy from treatment records.
\newblock \emph{Communications Medicine}, 4\penalty0 (1):\penalty0 245, 2024.

\bibitem[ChuGyouk(2024)]{kormedconceptsqa}
ChuGyouk.
\newblock Kormedconceptsqa.
\newblock \url{https://huggingface.co/datasets/ChuGyouk/KorMedConceptsQA}, 2024.

\bibitem[Cohan et~al.(2018)Cohan, Dernoncourt, Kim, Bui, Kim, Chang, and Goharian]{pubmedsummer}
Arman Cohan, Franck Dernoncourt, Doo~Soon Kim, Trung Bui, Seokhwan Kim, Walter Chang, and Nazli Goharian.
\newblock A discourse-aware attention model for abstractive summarization of long documents.
\newblock In \emph{Proceedings of the 2018 Conference of the North {A}merican Chapter of the Association for Computational Linguistics: Human Language Technologies, Volume 2 (Short Papers)}, pages 615--621, New Orleans, Louisiana, June 2018. Association for Computational Linguistics.
\newblock \doi{10.18653/v1/N18-2097}.
\newblock URL \url{https://aclanthology.org/N18-2097}.

\bibitem[Colicchio and Cimino(2019)]{39:coli:2019}
Tiago~K Colicchio and James~J Cimino.
\newblock Clinicians’ reasoning as reflected in electronic clinical note-entry and reading/retrieval: a systematic review and qualitative synthesis.
\newblock \emph{Journal of the American Medical Informatics Association}, 26\penalty0 (2):\penalty0 172--184, 2019.

\bibitem[ContactDoctor(2024)]{bmllama}
ContactDoctor.
\newblock Bio-medical: A high-performance biomedical language model.
\newblock https://huggingface.co/ContactDoctor/Bio-Medical-Llama-3-8B, 2024.

\bibitem[Evans et~al.(2021)Evans, Rhodes, Alhazzani, Antonelli, Coopersmith, French, Machado, Mcintyre, Ostermann, Prescott, et~al.]{fluid_volume_1}
Laura Evans, Andrew Rhodes, Waleed Alhazzani, Massimo Antonelli, Craig~M Coopersmith, Craig French, Fl{\'a}via~R Machado, Lauralyn Mcintyre, Marlies Ostermann, Hallie~C Prescott, et~al.
\newblock Surviving sepsis campaign: international guidelines for management of sepsis and septic shock 2021.
\newblock \emph{Critical care medicine}, 49\penalty0 (11):\penalty0 e1063--e1143, 2021.

\bibitem[Fang et~al.(2024)Fang, Liu, and Gong]{20:fang:2024}
Nan Fang, Guiliang Liu, and Wei Gong.
\newblock Offline inverse constrained reinforcement learning for safe-critical decision making in healthcare.
\newblock \emph{arXiv preprint arXiv:2410.07525}, 2024.

\bibitem[Fatemi et~al.(2022)Fatemi, Wu, Petch, Nelson, Connolly, Benz, Carnicelli, and Ghassemi]{26:fatemi:2022}
Mehdi Fatemi, Mary Wu, Jeremy Petch, Walter Nelson, Stuart~J Connolly, Alexander Benz, Anthony Carnicelli, and Marzyeh Ghassemi.
\newblock Semi-markov offline reinforcement learning for healthcare.
\newblock In \emph{Conference on Health, Inference, and Learning}, pages 119--137. PMLR, 2022.

\bibitem[Fujimoto et~al.(2019)Fujimoto, Meger, and Precup]{21:fjjimoto:2019}
Scott Fujimoto, David Meger, and Doina Precup.
\newblock Off-policy deep reinforcement learning without exploration.
\newblock In \emph{International conference on machine learning}, pages 2052--2062. PMLR, 2019.

\bibitem[Gordon(1999)]{43:gordon:1999}
Geoffrey~J Gordon.
\newblock \emph{Approximate solutions to Markov decision processes}.
\newblock Carnegie Mellon University, 1999.

\bibitem[Jayaraman et~al.(2024)Jayaraman, Desman, Sabounchi, Nadkarni, and Sakhuja]{RL_medicine_review}
Pushkala Jayaraman, Jacob Desman, Moein Sabounchi, Girish~N Nadkarni, and Ankit Sakhuja.
\newblock A primer on reinforcement learning in medicine for clinicians.
\newblock \emph{NPJ Digital Medicine}, 7\penalty0 (1):\penalty0 337, 2024.

\bibitem[Jiang and Li(2016)]{44:jiang:2016}
Nan Jiang and Lihong Li.
\newblock Doubly robust off-policy value evaluation for reinforcement learning.
\newblock In \emph{International conference on machine learning}, pages 652--661. PMLR, 2016.

\bibitem[Jin et~al.(2020)Jin, Pan, Oufattole, Weng, Fang, and Szolovits]{medqa}
Di~Jin, Eileen Pan, Nassim Oufattole, Wei-Hung Weng, Hanyi Fang, and Peter Szolovits.
\newblock What disease does this patient have? a large-scale open domain question answering dataset from medical exams, 2020.
\newblock URL \url{https://arxiv.org/abs/2009.13081}.

\bibitem[Jin et~al.(2019)Jin, Dhingra, Liu, Cohen, and Lu]{pubmedqa}
Qiao Jin, Bhuwan Dhingra, Zhengping Liu, William~W. Cohen, and Xinghua Lu.
\newblock Pubmedqa: A dataset for biomedical research question answering, 2019.
\newblock URL \url{https://arxiv.org/abs/1909.06146}.

\bibitem[Johnson et~al.(2016)Johnson, Pollard, Shen, Lehman, Feng, Ghassemi, Moody, Szolovits, Anthony~Celi, and Mark]{40:johnson:2016}
Alistair~EW Johnson, Tom~J Pollard, Lu~Shen, Li-wei~H Lehman, Mengling Feng, Mohammad Ghassemi, Benjamin Moody, Peter Szolovits, Leo Anthony~Celi, and Roger~G Mark.
\newblock Mimic-iii, a freely accessible critical care database.
\newblock \emph{Scientific data}, 3\penalty0 (1):\penalty0 1--9, 2016.

\bibitem[Johnson et~al.(2019)Johnson, Pollard, Berkowitz, Greenbaum, Lungren, Deng, Mark, and Horng]{mimic-cxr}
Alistair~EW Johnson, Tom~J Pollard, Seth~J Berkowitz, Nathaniel~R Greenbaum, Matthew~P Lungren, Chih-ying Deng, Roger~G Mark, and Steven Horng.
\newblock Mimic-cxr, a de-identified publicly available database of chest radiographs with free-text reports.
\newblock \emph{Scientific data}, 6\penalty0 (1):\penalty0 317, 2019.

\bibitem[Johnson et~al.(2023)Johnson, Bulgarelli, Shen, Gayles, Shammout, Horng, Pollard, Hao, Moody, Gow, et~al.]{41:johnson:2023}
Alistair~EW Johnson, Lucas Bulgarelli, Lu~Shen, Alvin Gayles, Ayad Shammout, Steven Horng, Tom~J Pollard, Sicheng Hao, Benjamin Moody, Brian Gow, et~al.
\newblock Mimic-iv, a freely accessible electronic health record dataset.
\newblock \emph{Scientific data}, 10\penalty0 (1):\penalty0 1, 2023.

\bibitem[Kaushik et~al.(2022)Kaushik, Kummetha, Moodley, and Bapi]{19:kaushik:2022}
Pramod Kaushik, Sneha Kummetha, Perusha Moodley, and Raju~S Bapi.
\newblock A conservative q-learning approach for handling distribution shift in sepsis treatment strategies.
\newblock \emph{arXiv preprint arXiv:2203.13884}, 2022.

\bibitem[Killian et~al.(2020)Killian, Zhang, Subramanian, Fatemi, and Ghassemi]{25:killian:2020}
Taylor~W Killian, Haoran Zhang, Jayakumar Subramanian, Mehdi Fatemi, and Marzyeh Ghassemi.
\newblock An empirical study of representation learning for reinforcement learning in healthcare.
\newblock \emph{arXiv preprint arXiv:2011.11235}, 2020.

\bibitem[Kiran et~al.(2021)Kiran, Sobh, Talpaert, Mannion, Al~Sallab, Yogamani, and P{\'e}rez]{20.5:kiran:2021}
B~Ravi Kiran, Ibrahim Sobh, Victor Talpaert, Patrick Mannion, Ahmad~A Al~Sallab, Senthil Yogamani, and Patrick P{\'e}rez.
\newblock Deep reinforcement learning for autonomous driving: A survey.
\newblock \emph{IEEE transactions on intelligent transportation systems}, 23\penalty0 (6):\penalty0 4909--4926, 2021.

\bibitem[Kocak et~al.(2012)Kocak, Bayrak, Erdamar, Ozparlak, Telatar, and Erogul]{32:kocak:2012}
Onur Kocak, Tuncay Bayrak, Aykut Erdamar, Levent Ozparlak, Ziya Telatar, and Osman Erogul.
\newblock Automated detection and classification of sleep apnea types using electrocardiogram (ecg) and electroencephalogram (eeg) features.
\newblock \emph{Advances in Electrocardiograms-Clinical Applications}, pages 211--230, 2012.

\bibitem[Komorowski et~al.(2018)Komorowski, Celi, Badawi, Gordon, and Faisal]{5:komor:2018}
Matthieu Komorowski, Leo~A Celi, Omar Badawi, Anthony~C Gordon, and A~Aldo Faisal.
\newblock The artificial intelligence clinician learns optimal treatment strategies for sepsis in intensive care.
\newblock \emph{Nature medicine}, 24\penalty0 (11):\penalty0 1716--1720, 2018.

\bibitem[Kostrikov et~al.(2021)Kostrikov, Nair, and Levine]{24:kostrikov:2021}
Ilya Kostrikov, Ashvin Nair, and Sergey Levine.
\newblock Offline reinforcement learning with implicit q-learning.
\newblock \emph{arXiv preprint arXiv:2110.06169}, 2021.

\bibitem[Kumar et~al.(2019)Kumar, Fu, Soh, Tucker, and Levine]{22:kumar:2019}
Aviral Kumar, Justin Fu, Matthew Soh, George Tucker, and Sergey Levine.
\newblock Stabilizing off-policy q-learning via bootstrapping error reduction.
\newblock \emph{Advances in neural information processing systems}, 32, 2019.

\bibitem[Kumar et~al.(2020)Kumar, Zhou, Tucker, and Levine]{23:kumar:2020}
Aviral Kumar, Aurick Zhou, George Tucker, and Sergey Levine.
\newblock Conservative q-learning for offline reinforcement learning.
\newblock \emph{Advances in neural information processing systems}, 33:\penalty0 1179--1191, 2020.

\bibitem[Kweon et~al.(2024)Kweon, Choi, Chu, Song, Hyeon, Gan, Kim, Kim, Park, and Choi]{kormedmcqa}
Sunjun Kweon, Byungjin Choi, Gyouk Chu, Junyeong Song, Daeun Hyeon, Sujin Gan, Jueon Kim, Minkyu Kim, Rae~Woong Park, and Edward Choi.
\newblock Kormedmcqa: Multi-choice question answering benchmark for korean healthcare professional licensing examinations, 2024.
\newblock URL \url{https://arxiv.org/abs/2403.01469}.

\bibitem[Lee et~al.(2024)Lee, Chung, Hyeon, Yang, Lee, Ryu, and Lee]{RL_medicine_1}
Hong~Yeul Lee, Soomin Chung, Dongwoo Hyeon, Hyun-Lim Yang, Hyung-Chul Lee, Ho~Geol Ryu, and Hyeonhoon Lee.
\newblock Reinforcement learning model for optimizing dexmedetomidine dosing to prevent delirium in critically ill patients.
\newblock \emph{npj Digital Medicine}, 7\penalty0 (1):\penalty0 325, 2024.

\bibitem[Lee et~al.(2023)Lee, Yoon, Kim, Park, Koo, Won, and Lee]{RL_medicine_2}
Hyeonhoon Lee, Hyun-Kyu Yoon, Jaewon Kim, Ji~Soo Park, Chang-Hoon Koo, Dongwook Won, and Hyung-Chul Lee.
\newblock Development and validation of a reinforcement learning model for ventilation control during emergence from general anesthesia.
\newblock \emph{npj Digital Medicine}, 6\penalty0 (1):\penalty0 145, 2023.

\bibitem[Levine et~al.(2020)Levine, Kumar, Tucker, and Fu]{36:levine:2020}
Sergey Levine, Aviral Kumar, George Tucker, and Justin Fu.
\newblock Offline reinforcement learning: Tutorial, review, and perspectives on open problems.
\newblock \emph{arXiv preprint arXiv:2005.01643}, 2020.

\bibitem[Li et~al.(2021)Li, Zhao, Lv, and Pan]{29:li:2021}
Yi~Li, Junli Zhao, Zhihan Lv, and Zhenkuan Pan.
\newblock Multimodal medical supervised image fusion method by cnn.
\newblock \emph{Frontiers in neuroscience}, 15:\penalty0 638976, 2021.

\bibitem[Li et~al.(2020)Li, Li, and Zhang]{sepsis_timing_1}
Yuting Li, Hongxiang Li, and Dong Zhang.
\newblock Timing of norepinephrine initiation in patients with septic shock: a systematic review and meta-analysis.
\newblock \emph{Critical Care}, 24:\penalty0 1--9, 2020.

\bibitem[Liang and Jia(2025)]{8:liang:2025}
Weijie Liang and Jinzhu Jia.
\newblock Reinforcement learning using neural networks in estimating an optimal dynamic treatment regime in patients with sepsis.
\newblock \emph{Computer Methods and Programs in Biomedicine}, page 108754, 2025.

\bibitem[Mahmood et~al.(2014)Mahmood, Van~Hasselt, and Sutton]{45:mahmood:2014}
A~Rupam Mahmood, Hado~P Van~Hasselt, and Richard~S Sutton.
\newblock Weighted importance sampling for off-policy learning with linear function approximation.
\newblock \emph{Advances in neural information processing systems}, 27, 2014.

\bibitem[Marik et~al.(2017)Marik, Linde-Zwirble, Bittner, Sahatjian, and Hansell]{fluid_volume_2}
Paul~E Marik, Walter~T Linde-Zwirble, Edward~A Bittner, Jennifer Sahatjian, and Douglas Hansell.
\newblock Fluid administration in severe sepsis and septic shock, patterns and outcomes: an analysis of a large national database.
\newblock \emph{Intensive care medicine}, 43:\penalty0 625--632, 2017.

\bibitem[Martin et~al.(2003)Martin, Mannino, Eaton, and Moss]{4:martin:2003}
Greg~S Martin, David~M Mannino, Stephanie Eaton, and Marc Moss.
\newblock The epidemiology of sepsis in the united states from 1979 through 2000.
\newblock \emph{New England Journal of Medicine}, 348\penalty0 (16):\penalty0 1546--1554, 2003.

\bibitem[Mathioudakis et~al.(2016)Mathioudakis, Rousalova, Gagnat, Saad, and Hardavella]{16:mathiou:2016}
Alexander Mathioudakis, Ilona Rousalova, Ane~Aamli Gagnat, Neil Saad, and Georgia Hardavella.
\newblock How to keep good clinical records.
\newblock \emph{Breathe}, 12\penalty0 (4):\penalty0 369--373, 2016.

\bibitem[Nambiar et~al.(2023)Nambiar, Ghosh, Ong, Chan, Bee, and Krishnaswamy]{27:nambiar:2023}
Mila Nambiar, Supriyo Ghosh, Priscilla Ong, Yu~En Chan, Yong~Mong Bee, and Pavitra Krishnaswamy.
\newblock Deep offline reinforcement learning for real-world treatment optimization applications.
\newblock In \emph{Proceedings of the 29th ACM SIGKDD conference on knowledge discovery and data mining}, pages 4673--4684, 2023.

\bibitem[Natarajan et~al.(2010)Natarajan, Stein, Jain, and Elhadad]{38:natar:2010}
Karthik Natarajan, Daniel Stein, Samat Jain, and No{\'e}mie Elhadad.
\newblock An analysis of clinical queries in an electronic health record search utility.
\newblock \emph{International journal of medical informatics}, 79\penalty0 (7):\penalty0 515--522, 2010.

\bibitem[Nauka et~al.(2025)Nauka, Kennedy, Brant, Komorowski, Pirracchio, Angus, and Seymour]{9:nauka:2025}
Peter~C Nauka, Jason~N Kennedy, Emily~B Brant, Matthieu Komorowski, Romain Pirracchio, Derek~C Angus, and Christopher~W Seymour.
\newblock Challenges with reinforcement learning model transportability for sepsis treatment in emergency care.
\newblock \emph{npj Digital Medicine}, 8\penalty0 (1):\penalty0 1--5, 2025.

\bibitem[Nie et~al.(2024)Nie, Chandak, Yuan, Badrinath, Flet-Berliac, and Brunskill]{42:nie:2024}
Allen Nie, Yash Chandak, Christina Yuan, Anirudhan Badrinath, Yannis Flet-Berliac, and Emma Brunskill.
\newblock Opera: Automatic offline policy evaluation with re-weighted aggregates of multiple estimators.
\newblock \emph{Advances in Neural Information Processing Systems}, 37:\penalty0 103652--103680, 2024.

\bibitem[Pal et~al.(2022)Pal, Umapathi, and Sankarasubbu]{medmcqa}
Ankit Pal, Logesh~Kumar Umapathi, and Malaikannan Sankarasubbu.
\newblock Medmcqa : A large-scale multi-subject multi-choice dataset for medical domain question answering, 2022.
\newblock URL \url{https://arxiv.org/abs/2203.14371}.

\bibitem[Paraskar et~al.(2025)Paraskar, Bhattacharya, and Kuttiappan]{34:paraskar:2025}
Gaurav Paraskar, Sankha Bhattacharya, and Anitha Kuttiappan.
\newblock The role of proteomics and genomics in the development of colorectal cancer diagnostic tools and potential new treatments.
\newblock \emph{ACS Pharmacology \& Translational Science}, 2025.

\bibitem[Raghavan et~al.(2014)Raghavan, Chen, Fosler-Lussier, and Lai]{15:raghavan:2014}
Preethi Raghavan, James~L Chen, Eric Fosler-Lussier, and Albert~M Lai.
\newblock How essential are unstructured clinical narratives and information fusion to clinical trial recruitment?
\newblock \emph{AMIA Summits on Translational Science Proceedings}, 2014:\penalty0 218, 2014.

\bibitem[Ruan et~al.(2025)Ruan, Tan, Ng, Huang, and Feng]{14:ruan:2025}
Yucheng Ruan, Daniel~J Tan, See~Kiong Ng, Ling Huang, and Mengling Feng.
\newblock Evidence-based multimodal fusion on structured ehrs and free-text notes for icu outcome prediction.
\newblock \emph{arXiv preprint arXiv:2501.04389}, 2025.

\bibitem[Russell(2006)]{3:Russell:2006}
James~A Russell.
\newblock Management of sepsis.
\newblock \emph{New England Journal of Medicine}, 355\penalty0 (16):\penalty0 1699--1713, 2006.

\bibitem[Shukla and Marlin(2020)]{13:shukla:2020}
Satya~Narayan Shukla and Benjamin~M Marlin.
\newblock Integrating physiological time series and clinical notes with deep learning for improved icu mortality prediction.
\newblock \emph{arXiv preprint arXiv:2003.11059}, 2020.

\bibitem[Singer et~al.(2016)Singer, Deutschman, Seymour, Shankar-Hari, Annane, Bauer, Bellomo, Bernard, Chiche, Coopersmith, et~al.]{1:singer:2016}
Mervyn Singer, Clifford~S Deutschman, Christopher~Warren Seymour, Manu Shankar-Hari, Djillali Annane, Michael Bauer, Rinaldo Bellomo, Gordon~R Bernard, Jean-Daniel Chiche, Craig~M Coopersmith, et~al.
\newblock The third international consensus definitions for sepsis and septic shock (sepsis-3).
\newblock \emph{Jama}, 315\penalty0 (8):\penalty0 801--810, 2016.

\bibitem[Singh et~al.(2022)Singh, Kumar, and Singh]{20.7:singh:2022}
Bharat Singh, Rajesh Kumar, and Vinay~Pratap Singh.
\newblock Reinforcement learning in robotic applications: a comprehensive survey.
\newblock \emph{Artificial Intelligence Review}, 55\penalty0 (2):\penalty0 945--990, 2022.

\bibitem[Song et~al.(2021)Song, Zheng, Li, Lu, Zhu, and Shen]{28:song:2021}
Juan Song, Jian Zheng, Ping Li, Xiaoyuan Lu, Guangming Zhu, and Peiyi Shen.
\newblock An effective multimodal image fusion method using mri and pet for alzheimer's disease diagnosis.
\newblock \emph{Frontiers in digital health}, 3:\penalty0 637386, 2021.

\bibitem[Sutton et~al.(1998)Sutton, Barto, et~al.]{35:sutton:1998}
Richard~S Sutton, Andrew~G Barto, et~al.
\newblock \emph{Reinforcement learning: An introduction}, volume~1.
\newblock MIT press Cambridge, 1998.

\bibitem[Team(2025{\natexlab{a}})]{gemma3}
Gemma Team.
\newblock Gemma 3.
\newblock 2025{\natexlab{a}}.
\newblock URL \url{https://goo.gle/Gemma3Report}.

\bibitem[Team(2025{\natexlab{b}})]{qwen3}
Qwen Team.
\newblock Qwen3 technical report, 2025{\natexlab{b}}.
\newblock URL \url{https://arxiv.org/abs/2505.09388}.

\bibitem[Teles et~al.(2025)Teles, de~Moura, Silva, Roberts, and Stahl]{17:teles:2025}
Ariel~Soares Teles, Ivan~Rodrigues de~Moura, Francisco Silva, Angus Roberts, and Daniel Stahl.
\newblock Ehr-based prediction modelling meets multimodal deep learning: A systematic review of structured and textual data fusion methods.
\newblock \emph{Information Fusion}, page 102981, 2025.

\bibitem[Tolstikhin et~al.(2021)Tolstikhin, Houlsby, Kolesnikov, Beyer, Zhai, Unterthiner, Yung, Steiner, Keysers, Uszkoreit, Lucic, and Dosovitskiy]{tol:ranzato:2021}
Ilya~O Tolstikhin, Neil Houlsby, Alexander Kolesnikov, Lucas Beyer, Xiaohua Zhai, Thomas Unterthiner, Jessica Yung, Andreas Steiner, Daniel Keysers, Jakob Uszkoreit, Mario Lucic, and Alexey Dosovitskiy.
\newblock Mlp-mixer: An all-mlp architecture for vision.
\newblock In M.~Ranzato, A.~Beygelzimer, Y.~Dauphin, P.S. Liang, and J.~Wortman Vaughan, editors, \emph{Advances in Neural Information Processing Systems}, volume~34, pages 24261--24272. Curran Associates, Inc., 2021.
\newblock URL \url{https://proceedings.neurips.cc/paper_files/paper/2021/file/cba0a4ee5ccd02fda0fe3f9a3e7b89fe-Paper.pdf}.

\bibitem[Touvron et~al.(2023)Touvron, Lavril, Izacard, Martinet, Lachaux, Lacroix, Rozière, Goyal, Hambro, Azhar, Rodriguez, Joulin, Grave, and Lample]{llama}
Hugo Touvron, Thibaut Lavril, Gautier Izacard, Xavier Martinet, Marie-Anne Lachaux, Timothée Lacroix, Baptiste Rozière, Naman Goyal, Eric Hambro, Faisal Azhar, Aurelien Rodriguez, Armand Joulin, Edouard Grave, and Guillaume Lample.
\newblock Llama: Open and efficient foundation language models, 2023.
\newblock URL \url{https://arxiv.org/abs/2302.13971}.

\bibitem[Tu et~al.(2025)Tu, Luo, Pan, Wang, Su, Zhang, and Wang]{18:tu:2025}
Rui Tu, Zhipeng Luo, Chuanliang Pan, Zhong Wang, Jie Su, Yu~Zhang, and Yifan Wang.
\newblock Offline safe reinforcement learning for sepsis treatment: Tackling variable-length episodes with sparse rewards.
\newblock \emph{Human-Centric Intelligent Systems}, 5\penalty0 (1):\penalty0 63--76, 2025.

\bibitem[Vincent et~al.(2014)Vincent, Marshall, {\~N}amendys-Silva, Fran{\c{c}}ois, Martin-Loeches, Lipman, Reinhart, Antonelli, Pickkers, Njimi, et~al.]{2:vincent:2014}
Jean-Louis Vincent, John~C Marshall, Silvio~A {\~N}amendys-Silva, Bruno Fran{\c{c}}ois, Ignacio Martin-Loeches, Jeffrey Lipman, Konrad Reinhart, Massimo Antonelli, Peter Pickkers, Hassane Njimi, et~al.
\newblock Assessment of the worldwide burden of critical illness: the intensive care over nations (icon) audit.
\newblock \emph{The lancet Respiratory medicine}, 2\penalty0 (5):\penalty0 380--386, 2014.

\bibitem[Waechter et~al.(2014)Waechter, Kumar, Lapinsky, Marshall, Dodek, Arabi, Parrillo, Dellinger, Garland, of~Septic Shock Database Research~Group, et~al.]{sepsis_timing_2}
Jason Waechter, Anand Kumar, Stephen~E Lapinsky, John Marshall, Peter Dodek, Yaseen Arabi, Joseph~E Parrillo, R~Phillip Dellinger, Allan Garland, Cooperative Antimicrobial~Therapy of~Septic Shock Database Research~Group, et~al.
\newblock Interaction between fluids and vasoactive agents on mortality in septic shock: a multicenter, observational study.
\newblock \emph{Critical care medicine}, 42\penalty0 (10):\penalty0 2158--2168, 2014.

\bibitem[Wang et~al.(2024)Wang, Liu, Yang, Wang, Xiong, Cheng, and Wu]{7:wang:2024}
Yuan Wang, Anqi Liu, Jucheng Yang, Lin Wang, Ning Xiong, Yisong Cheng, and Qin Wu.
\newblock Clinical knowledge-guided deep reinforcement learning for sepsis antibiotic dosing recommendations.
\newblock \emph{Artificial Intelligence in Medicine}, 150:\penalty0 102811, 2024.

\bibitem[Wang et~al.(2016)Wang, Schaul, Hessel, Hasselt, Lanctot, and Freitas]{ddqn}
Ziyu Wang, Tom Schaul, Matteo Hessel, Hado Hasselt, Marc Lanctot, and Nando Freitas.
\newblock Dueling network architectures for deep reinforcement learning.
\newblock In \emph{International conference on machine learning}, pages 1995--2003. PMLR, 2016.

\bibitem[Watkins and Dayan(1992)]{37:watkins:1992}
Christopher~JCH Watkins and Peter Dayan.
\newblock Q-learning.
\newblock \emph{Machine learning}, 8:\penalty0 279--292, 1992.

\bibitem[Xie et~al.(2024)Xie, Chen, Chen, Peng, Hu, Lin, Peng, Huang, Zhang, Keloth, He, Ohno-Machido, Wu, Xu, and Bian]{medllama}
Qianqian Xie, Qingyu Chen, Aokun Chen, Cheng Peng, Yan Hu, Fongci Lin, Xueqing Peng, Jimin Huang, Jeffrey Zhang, Vipina Keloth, Huan He, Lucila Ohno-Machido, Yonghui Wu, Hua Xu, and Jiang Bian.
\newblock Me llama: Foundation large language models for medical applications, 2024.

\bibitem[Yeo et~al.(2022)Yeo, Lee, Kim, Jang, Lee, Oh, Park, Lim, Cho, et~al.]{sepsis_timing_3}
Hye~Ju Yeo, Young~Seok Lee, Tae~Hwa Kim, Jin~Ho Jang, Heung~Bum Lee, Dong~Kyu Oh, Mi~Hyeon Park, Chae-Man Lim, Woo~Hyun Cho, et~al.
\newblock Vasopressor initiation within 1 hour of fluid loading is associated with increased mortality in septic shock patients: analysis of national registry data.
\newblock \emph{Critical Care Medicine}, 50\penalty0 (4):\penalty0 e351--e360, 2022.

\bibitem[Zhang et~al.(2023)Zhang, Xue, Li, Zou, and Zhu]{30:zhang:2023}
Menghao Zhang, Minghao Xue, Shuying Li, Yun Zou, and Quing Zhu.
\newblock Fusion deep learning approach combining diffuse optical tomography and ultrasound for improving breast cancer classification.
\newblock \emph{Biomedical Optics Express}, 14\penalty0 (4):\penalty0 1636--1646, 2023.

\bibitem[Zhu et~al.(2024)Zhu, Wang, He, Xie, Zheng, Ma, and Pan]{11:zhu:2024}
Yinghao Zhu, Zixiang Wang, Long He, Shiyun Xie, Xiaochen Zheng, Liantao Ma, and Chengwei Pan.
\newblock Prism: Mitigating ehr data sparsity via learning from missing feature calibrated prototype patient representations.
\newblock In \emph{Proceedings of the 33rd ACM International Conference on Information and Knowledge Management}, pages 3560--3569, 2024.

\end{thebibliography}

\newpage

\appendix

\section{Data availability}
The MIMIC datasets are publicly available through Physionet (https://physionet.org/about/database). 
Due to the ethical restrictions imposed by the IRB, the private dataset is only available upon reasonable request. The source code of the MORE-CLEAR framework can be found at : https://anonymous.4open.science/r/MORE-CLEAR-FD32

\section{LLM Performance in Note Summarization}
The clinical notes contain a mixture of medical jargon in English and other languages.
In order to assess the efficacy of LLM models in a bilingual setting, evaluations on question answering and summarization were conducted (Table \ref{tab:LLM_benchmark}).
The Gemma3-27B-it performed well overall, thus it was employed for our note summarization.

\begin{table*}[ht!]
  \centering
  \footnotesize
  \setlength{\tabcolsep}{4pt}
  \resizebox{\textwidth}{!}{%
    \begin{tabular}{l
      cc cc cc cc cc
      cc cc}
      \toprule
      & \multicolumn{10}{c}{\textbf{Question Answering}}

      & \multicolumn{4}{c}{\textbf{Text Summarization}} \\
      \cmidrule(lr){2-11} \cmidrule(lr){12-15}
        \textbf{LLMs}
        & \multicolumn{2}{c}{PubMedQA\cite{pubmedqa}}
        & \multicolumn{2}{c}{MedMCQA\cite{medmcqa}}
        & \multicolumn{2}{c}{MedQA\cite{medqa}}
        & \multicolumn{2}{c}{KorMedMCQA\cite{kormedmcqa}}
        & \multicolumn{2}{c}{KorMedConceptsQA\cite{kormedconceptsqa}}
        & \multicolumn{2}{c}{Pubmed\cite{pubmedsummer}}
        & \multicolumn{2}{c}{MIMIC-CXR\cite{mimic-cxr}} \\
      \cmidrule(lr){2-3}\cmidrule(lr){4-5}\cmidrule(lr){6-7}\cmidrule(lr){8-9}\cmidrule(lr){10-11}\cmidrule(lr){12-13}\cmidrule(lr){14-15}
      & Acc & Macro-F1
      & Acc & Macro-F1
      & Acc & Macro-F1
      & Acc & Macro-F1
      & Acc & Macro-F1
      & R-1 & BertS
      & R-1 & BertS \\
      \midrule
      Llama3.2-3B-it \cite{llama}
        & \textbf{0.734} & \textbf{0.514}
        & 0.522 & 0.514
        & 0.542 & 0.539
        & 0.359 & 0.354
        & 0.299 & 0.271
        & \textbf{0.381} & \textbf{0.837}
        & 0.068 & 0.816 \\
      Llama3.1-8B-it \cite{llama}
        & \textbf{0.758} & \textbf{0.531}
        & \textbf{0.570} & \textbf{0.569}
        & \textbf{0.611} & \textbf{0.608}
        & \textbf{0.484} & \textbf{0.479}
        & \textbf{0.561} & \textbf{0.555}
        & \textbf{0.384} & 0.836
        & 0.078 & 0.825 \\
      Med-Llama3-8B \cite{medllama}
        & 0.660 & 0.332
        & 0.480 & 0.479
        & 0.563 & 0.281
        & 0.198 & 0.064
        & 0.538 & 0.531
        & 0.335 & \textbf{0.825}
        & \textbf{0.213} & \textbf{0.854} \\
      Bio-Medical Llama3-8B \cite{bmllama}
        & 0.478 & 0.346
        & 0.333 & 0.320
        & 0.426 & 0.421
        & 0.200 & 0.084
        & 0.267 & 0.245
        & 0.325 & 0.826
        & 0.080 & 0.828 \\
      Gemma3-27B-it \cite{gemma3}
        & 0.474 & 0.423
        & \textbf{0.560} & \textbf{0.557}
        & \textbf{0.685} & \textbf{0.679}
        & \textbf{0.660} & \textbf{0.662}
        & 0.342 & 0.303
        & 0.363 & \textbf{0.840}
        & \textbf{0.179} & \textbf{0.855} \\
      Qwen3-32B \cite{qwen3}
        & 0.454 & 0.309
        & 0.297 & 0.296
        & 0.637 & 0.454
        & 0.318 & 0.271
        & \textbf{0.572} & 0.469
        & 0.355 & 0.830
        & 0.091 & 0.834 \\
      \bottomrule
    \end{tabular}%
    }
  \caption{LLM Evaluation on Medical QA and Summarization Benchmarks. R-1: Rouge-1, BertS: BertScore}
  \label{tab:LLM_benchmark}
\end{table*}

\section{Window size for note stack}
Table \ref{tab:stack-window} shows an ablation of context-window size on OPE in three cohorts. 
When the window size \(W\) is 3, the highest OPERA scores and DR estimates are achieved, indicating the highest expected return and minimal bias. 
As \(W\) increases, OPERA and DR decline monotonically, suggesting that compact contextual windows more effectively guide policy optimization.
This tendency appears consistently in both FQE and WIS.
However, WIS remains invariant comparably, implying that importance‐weight variance is largely unaffected by window length.

\begin{table}[ht]
  \centering
  \captionsetup{skip=10pt}
  \scriptsize     
    \begin{tabular}{llccc}
      \toprule
       &  & \multicolumn{3}{c}{\textbf{Window Size}} \\
      \cmidrule(lr){3-5}
      \textbf{Dataset} & \textbf{Metric}
        & \textbf{W=3} & \textbf{W=5} & \textbf{W=7} \\
      \midrule
      \multirow{4}{*}{MIMIC-III}
        & $\uparrow$ OPERA & 2.705 ± 0.169 & 2.551 ± 0.154 & 2.316 ± 0.064 \\
        & $\uparrow$ DR    & 2.412 ± 0.038 & 2.161 ± 0.124 & 1.906 ± 0.088 \\
        & $\uparrow$ FQE   & 1.058 ± 0.489 & 1.666 ± 0.544 & 0.795 ± 0.208 \\
        & $\uparrow$ WIS   & 0.672 ± 0.016 & 0.653 ± 0.024 & 0.630 ± 0.031 \\
      \midrule
      \multirow{4}{*}{MIMIC-IV}
        & $\uparrow$ OPERA & 3.581 ± 0.233 & 3.478 ± 0.173 & 3.002 ± 0.220 \\
        & $\uparrow$ DR    & 3.375 ± 0.126 & 3.302 ± 0.113 & 2.757 ± 0.182 \\
        & $\uparrow$ FQE   & 1.711 ± 0.491 & 1.389 ± 0.261 & 1.363 ± 0.393 \\
        & $\uparrow$ WIS   & 0.794 ± 0.023 & 0.793 ± 0.035 & 0.772 ± 0.033 \\
      \midrule
      \multirow{4}{*}{Private Dataset}
        & $\uparrow$ OPERA & 2.403 ± 0.190 & 1.986 ± 0.079 & 1.465 ± 0.137 \\
        & $\uparrow$ DR    & 2.088 ± 0.110 & 1.573 ± 0.061 & 1.112 ± 0.066 \\
        & $\uparrow$ FQE   & 6.495 ± 5,347 & 3.910 ± 1.349 & 3.141 ± 1.468 \\
        & $\uparrow$ WIS   & 0.721 ± 0.028 & 0.697 ± 0.039 & 0.719 ± 0.040 \\
      \bottomrule
    \end{tabular}%
  \caption{Effect of stacking window size on performance}
  \label{tab:stack-window}
\end{table}

\end{document}